%% file: n_dude_theory_arxiv.tex
\documentclass{article}
\input{taesup.sty}

\usepackage{dsfont}

\newcommand{\eg}{{\it e.g.}}
\newcommand{\ie}{{\it i.e.}}

\newcommand{\Pib}{\mathbf{\Pi}}
\newcommand{\Lb}{\mathbf{\Lambda}}
\newcommand{\Ellb}{\mathbf{L}}

\usepackage{microtype}
\usepackage{graphicx}
\usepackage{subfigure}
\usepackage{booktabs} 

\usepackage{hyperref}

\usepackage{wrapfig}
\usepackage{multirow}

\usepackage[accepted]{icml2018}

\icmltitlerunning{A Denoising Loss Bound for Neural Network based Universal Discrete Denoisers}

\begin{document}

\twocolumn[
\icmltitle{A Denoising Loss Bound for Neural Network based \\Universal Discrete Denoisers}



\icmlsetsymbol{equal}{*}

\begin{icmlauthorlist}
\icmlauthor{Taesup Moon}{to}
\end{icmlauthorlist}

\icmlaffiliation{to}{Department of Electronic and Electrical Engineering, Sungkyunkwan University (SKKU), Suwon, Korea}

\icmlcorrespondingauthor{Taesup Moon}{tsmoon@skku.edu}

\icmlkeywords{Machine Learning, ICML}

\vskip 0.3in
]




\begin{abstract}
We obtain a denoising loss bound of the recently proposed neural network based universal discrete denoiser, Neural DUDE, which can adaptively learn its parameters solely from the noise-corrupted data, by minimizing the \emph{empirical estimated loss}. The resulting bound resembles the generalization error bound of the standard empirical risk minimizers (ERM) in supervised learning, and we show that the well-known bias-variance tradeoff also exists in our loss bound. The key tool we develop is the concentration of the unbiased estimated loss on the true denoising loss, which is shown to hold \emph{uniformly} for \emph{all} bounded network parameters and \emph{all} underlying clean sequences. For proving our main results, we make a novel application of the tools from the statistical learning theory. Finally, we show that the hyperparameters of Neural DUDE can be chosen from a small validation set to significantly improve the denoising performance, as predicted by the theoretical result of this paper. 

\end{abstract}

\input{intro_arxiv.tex}
\input{notation_arxiv.tex}

\input{main_arxiv.tex}

\input{proof_arxiv.tex}
\input{experiments_arxiv.tex}
\bibliographystyle{icml2018}

\clearpage
 \input{appendix}




\end{document}

%% file: intro_arxiv.tex

\section{Introduction}

Cleaning noise-corrupted discrete-valued data, also known as discrete denoising, has diverse application areas, \eg, image denoising and DNA sequence denoising. \emph{Universal} discrete denoising, which assumes nothing about the data except for the noise mechanism being a discrete memoryless channel (DMC), has been first considered in \citep{weissman2005universal}. Discrete Universal DEnoiser (DUDE), which operates as a sliding-window denoiser, proposed in \citep{weissman2005universal} enjoyed the rigorous theoretical performance guarantees in the universal setting as well as state-of-the-art empirical performance for several applications, \eg, binary image denoising \citep{Ordetal03} and DNA sequence denoising \citep{LeeMooYooWei16}. 

Despite the strong results of DUDE, one of its major drawbacks was that the performance of the method is highly sensitive on the choice of its hyperparameter, \ie, the sliding-window size $k$. In order to overcome this drawback, which is mainly due to the nonparamtric nature of DUDE that separately obtains the empirical counts of the noisy symbols for different contexts, \citep{MooMinLeeYoo16} recently introduced a neural network based sliding-window denoiser, dubbed as Neural DUDE. By devising novel ``pseudo-labels'', Neural DUDE was able to \emph{learn} a single neural network-based sliding window denoiser, solely from the noisy data, that shares information from similar contexts through its parameters. As a result, the experiments in \citep{MooMinLeeYoo16} showed that Neural DUDE can significantly outperform DUDE and maintain robustness with respect to $k$.

While the approach of Neural DUDE has been recently extended to the case of continuous-valued signals as well \citep{ChaMoo18}, a theoretical analyses on the performance of Neural DUDE has been lacking. In this paper, we leverage the tools from statistical learning theory, \eg, Rademacher complexity \citep{BarMen02,cs229T}, and obtain a denoising loss bound for the class of neural network-based sliding window denoisers that are learned by minimizing the \emph{empirical estimated loss}, such as Neural DUDE. Such bound closely resembles the generalization bound for the standard empirical risk minimizers (ERM) in supervised learning. 
The main difference is that while the supervised learning cares about the prediction performance for the unseen test data, we care about the denoising performance for the \emph{unseen underlying clean $x^n$}. 
The key tool we develop for showing our main result is the concentration of the average estimated loss (computed from noisy data only), which are not the average of independent random variables unlike in the standard ERM, on the true denoising loss (computed from clean and noisy data). 
The concentration is shown to hold in a strong sense; namely, it holds with high probability \emph{uniformly} over \emph{all} neural network based denoisers that have the same architecture as Neural DUDE and for \emph{all} underlying clean sequences, when an appropriate condition holds.

The resulting denoising loss bound can be interpreted with the standard bias-variance tradeoff in supervised learning. Therefore, the theory guides to choose the hyperparameters of Neural DUDE, such as the context size $k$, the number of layers, hidden nodes and epochs, from a separate validation set that consists of small number of data that has similar charateristics with the given data subject to denoising. 
As a result, we show in our experiment that the hyperparameters chosen from a separate validation set can lead to much better denoising results than simple hand-picked ones in the original \citep{MooMinLeeYoo16}.

The rest of the paper is organized as follows. After reviewing some necessary notations and prelimiaries on Neural DUDE in Section \ref{sec:notation_prelim}, the main results of the paper is stated in Section \ref{sec:main results}. Section \ref{sec:proof} presents three lemmas and the proof of the main theorems. The proof of lemmas are deferred to the Supplementary Material. Section \ref{sec:experiments} gives experimental results that corroborate our theoretical finding, and Section \ref{sec:conclusion} concludes with some future directions. 

%% file: notation_arxiv.tex
\section{Notations and Preliminaries}\label{sec:notation_prelim}

To be self-contained, we introduce notations that mainly follow \citep{MooMinLeeYoo16}. Throughout the paper, an $n$-tuple sequence is denoted as, \eg, $a^n=(a_1,\ldots,a_n)$, and $a_i^j$ refers to the subsequence $(a_i,\ldots,a_j)$. The uppercase letters will stand for the random variables, and the lowercase letters will stand for either the realizations of the random variables or the individual symbols. We denote $\Delta^d$ as the probability simplex in $\mathbb{R}^d$. In \emph{universal} discrete denoising, the clean, underlying source data will be denoted as an \emph{individual sequence} $x^n$ as we do not assume any probabilistic models on it. We assume each component $x_i$ takes a value in some finite set $\mcX$. For example, for binary data, $\mcX=\{0,1\}$, and for DNA data, $\mcX=\{\texttt{A},\texttt{C},\texttt{G},\texttt{T}\}$. 

When the source sequence is corrupted by a Discrete Memoryless Channel (DMC), namely, the index-independent noise, it results in a noisy version of the source, $Z^n$, of which each $Z_i$ takes a value in, again, a finite set $\mcZ$.
The DMC is completely characterized by the channel transition matrix $\mathbf\Pi\in\mathbb{R}^{|\mcX|\times|\mcZ|}$, of which the $(x,z)$-th element stands for $\text{Pr}(Z=z|X=x)$, \ie, the conditional probability of the noisy symbol taking value $z$ given the source symbol was $x$. An essential but natural assumption we make is that $\mathbf{\Pi}$ is of the \emph{full row rank}. We also denote $\Pib^\dagger=\Pib^\top(\Pib\Pib^\top)^{-1}$ as the Moore-Penrose pseudoinverse of $\Pib$. In our setting, $\Pib$ is assumed to be known to the denoiser. 

Upon observing the entire noisy data $Z^n$, a discrete denoiser reconstructs the original data with $\hat{X}^n=(\hat{X}_1(Z^n),\ldots,\hat{X}_n(Z^n))$, where each reconstructed symbol $\hat{X}_i(Z^n)$ takes its value in a finite set $\hat{\mathcal{X}}$. The goodness of the reconstruction is measured by the average denoising loss,
$
\frac{1}{n}\sum_{i=1}^n\Lb(x_i,\hat{X}_i(Z^n)),
$
where $\Lb(x_i,\hat{x}_i)$ is a loss function that measures the loss incurred by estimating $x_i$ with $\hat{x}_i$. The loss function is fully represented with a loss matrix $\mathbf{\Lambda}\in\mathbb{R}^{|\mcX|\times|\hat{\mcX}|}$. 

The $k$-th order sliding window denoisers are the denoisers that are defined by a time-invariant mapping $s_k:\mcZ^{2k+1}\rightarrow\hat{\mcX}$. That is, $\hat{X}_i(Z^n)= s_k(Z_{i-k}^{i+k})$. We also denote the tuple $(Z_{i-k}^{i-1},Z_{i+1}^{i+k})\triangleq\Cb_i$ as the $k$-th order double-sided context\footnote{Note we are using the uppercase notation $\Cb_i$ to highlight the randomness as opposed to $\cb_i$ used in \citep{MooMinLeeYoo16}.} around the noisy symbol $Z_i$, and we let $\mathbf{C}[k]$ as the set of all such contexts. As discussed in \citep{MooMinLeeYoo16}, both DUDE in \citep{weissman2005universal} and Neural DUDE are sliding window denoisers. We also denote $\mcS\triangleq\{s:\mcZ\rightarrow\hat{\mcX}\}$ as the set of \emph{single-symbol denoisers}  that are sliding window denoisers with $k=0$. Note $|\mcS|=|\hat{\mcX}|^{|\mcZ|}$. Then, an alternative view of of $s_k(\cdot)$ is that $s_k(\Cb_i,\cdot)\in\mcS$ is a single symbol denoiser defined by $\Cb_i$ and applied to $Z_i$. 



The basic building block of Neural DUDE is the  \emph{unbiased estimated} loss function as described in \citep[Section 3.1]{MooMinLeeYoo16}. That is, based on the known $\Pib$ assumption, we can devise an esimated loss 
\be
\Ellb=\Pib^\dagger\bm{\rho}\in\mathbb{R}^{|\mcZ|\times|\mcS|},\label{eq:est_loss}
\ee in which $\bm{\rho}\in\mathbb{R}^{|\mcX|\times|\mcS|}$ with the $(x,s)$-th element is $\mathbb{E}_{Z|x}\Lb(x,s(Z))$. The notation $\mathbb{E}_{Z|x}(\cdot)$ stands for the expectation with respect to the distribution Pr$(Z=\cdot|X=x)$ defined by the $x$-th row of $\Pib$. Then, as shown in \citep{MooMinLeeYoo16,UFP06}, $\Ellb$ has the unbiased property, $\mathbb{E}_{Z|x}\Ellb(Z,s)=\mathbb{E}_{Z|x}\Lb(x,s(Z))$.

\subsection{Neural DUDE}

\begin{figure*}[h]
    \centering
    \subfigure[An example architecture]{\label{fig:n_dude_arch}
    \includegraphics[width=0.3\textwidth]{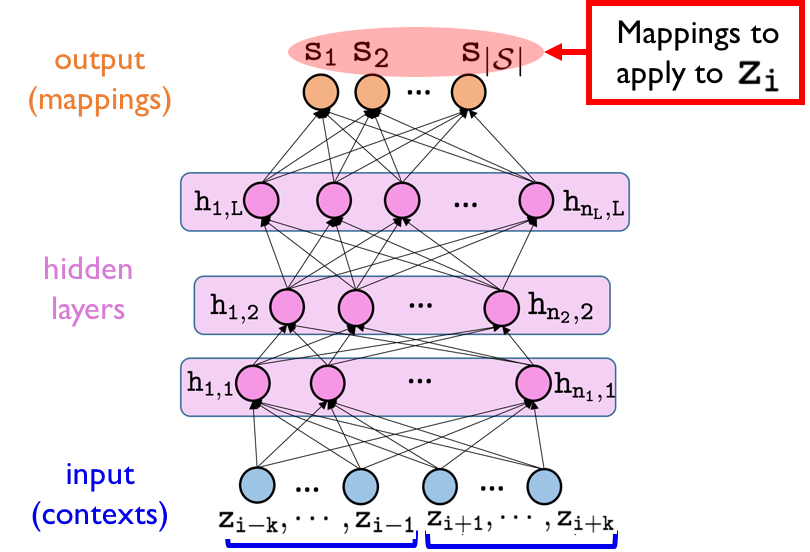}}
    \subfigure[BER plot for DUDE]{\label{fig:dude_est_ber}
    \includegraphics[width=0.3\textwidth]{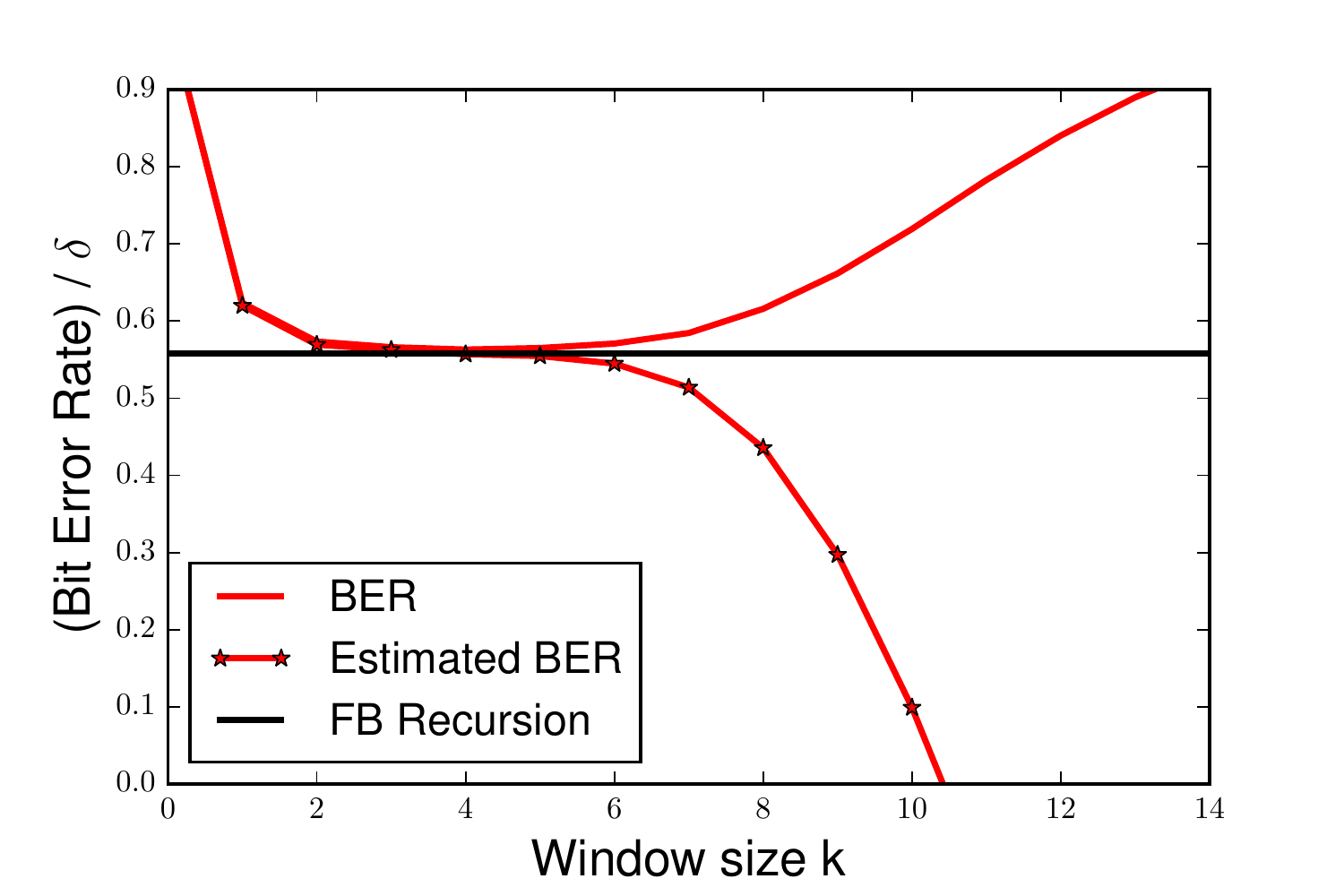}}
    \subfigure[BER plot for Neural DUDE]{\label{fig:n_dude_est_ber}
    \includegraphics[width=0.3\textwidth]{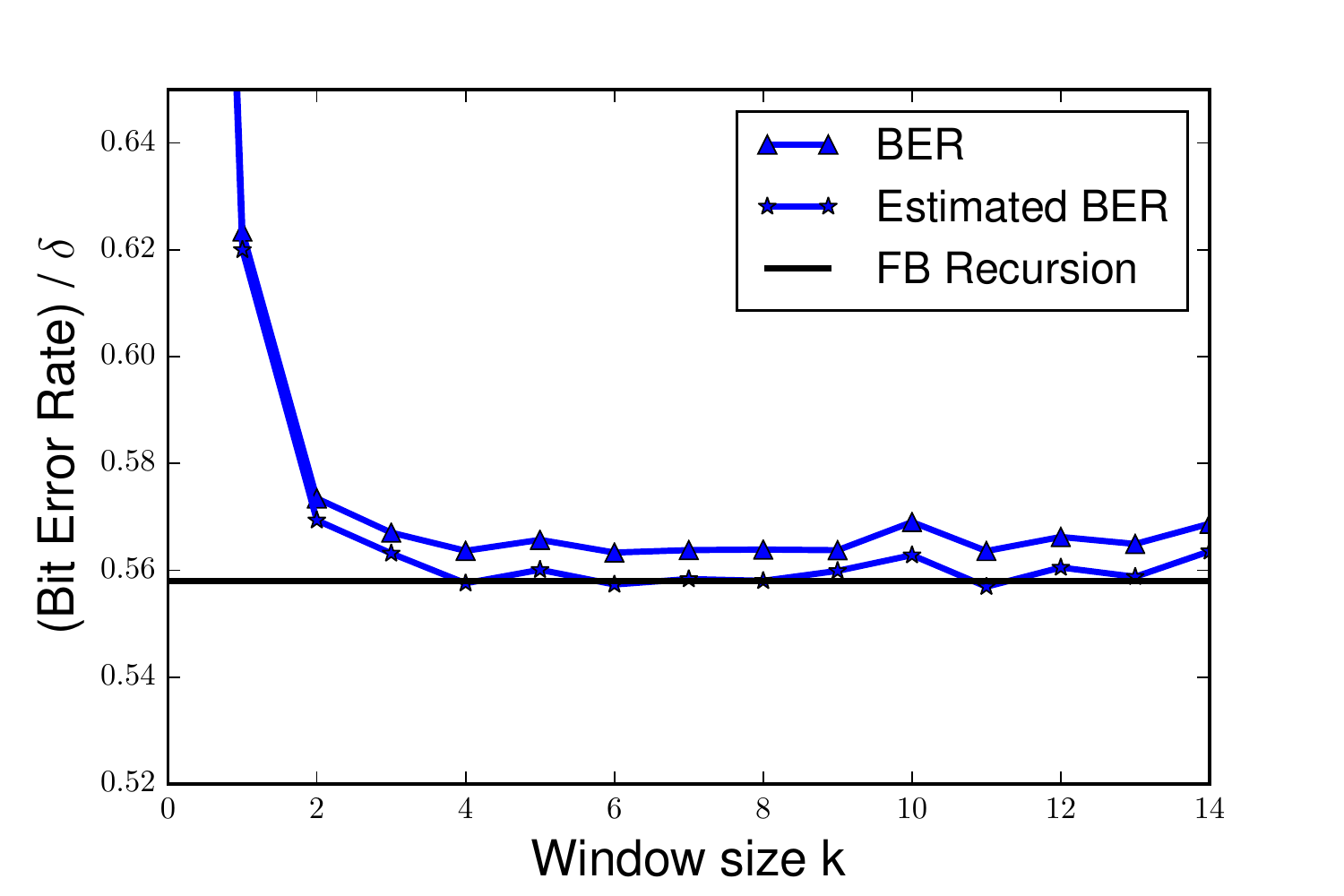}}
    \caption{(a) An example network architecture $\mathbf{p}^k(\wb,\cdot)$ of Neural DUDE with $L$ layers. (b),(c) The Bit Error Rate (BER) and the estimated BER plots for DUDE and Neural DUDE for the synthetic binary example in \citep[Section 5.1]{MooMinLeeYoo16}. 
    }
    \label{fig:n_dude}
\end{figure*}

Neural DUDE \citep{MooMinLeeYoo16} defines a \emph{single} fully-connected neural network $\mathbf{p}^k(\mathbf{w},\cdot):\mathcal{Z}^{2k}\rightarrow\Delta^{|\mcS|}$ that works as a sliding-window denoiser. 
That is, at location $i$, the network takes the double-sided context $\Cb_i\in\Cb[k]$ as input and outputs the probability distribution on the single symbol denoisers to apply to $Z_i$. We let $\mathbf{w}$ stand for all the parameters in the network. 
Figure \ref{fig:n_dude_arch} shows an example architecture of Neural DUDE in which $L$ is the total number of layers and $n_\ell$ is the number of nodes in the $\ell$-th layer. By carrying out the one-hot encoding of each noisy symbol, the total number of parameters of Neural DUDE then becomes $2k|\mcZ|n_1+\sum_{\ell=2}^Ln_{\ell-1}n_{\ell}+n_L|\mcS|$. Furthermore, the ReLU (Rectified Linear Unit), $f(x)=\max\{0,x\}$, is used as the activation function for all intermediate hidden nodes. For the output layer, the usual softmax function is used. 

In order to train the network parameters, Neural DUDE computes the matrix $\Ellb_{\text{new}}\in\mathbb{R}^{|\mcZ|\times|\mcS|}$ defined as 
\be
\Ellb_{\text{new}} \triangleq -\Ellb+L_{\text{max}}\mathbf{1}_{|\mcZ|}\mathbf{1}_{|\mcS|}^\top,\label{eq:L_max}
\ee
in which $L_{\text{max}}\triangleq \max_{z,s}\Ellb(z,s)$, and $\mathbf{1}_{|\mcZ|}$ and $\mathbf{1}_{|\mcS|}$ stand for the all-1 vector with $|\mcZ|$ and $|\mcS|$ dimensions, respectively. 
Note that all the elements in $\Ellb_{\text{new}}$ can be computed with $z$ and $s$ (and \emph{not} with $x$) and are designed to be non-negative. 
Once $\Ellb_{\text{new}}$ is computed, Neural DUDE uses the objective function in \citep[Eq.(7)]{MooMinLeeYoo16}, 
\be
\mathcal{L}(\mathbf{w}, Z^n)&\triangleq&\frac{1}{n}\sum_{i=1}^n\mathcal{C}\Big(\mathbf{L}_{\text{new}}^\top\mathds{1}_{Z_i}, \mathbf{p}^k(\mathbf{w},\mathbf{C}_i)\Big),\label{eq:objective}
\ee
in which $\mathcal{C}(\mathbf{g},\mathbf{p})=-\sum_{i=1}^{|\mcS|} g_i\log p_i$ for $\mathbf{g}\in\mathbb{R}_{+}^{|\mcS|}$ and $\mathbf{p}\in\Delta^{|\mcS|}$ stands for the (unnormalized) cross-entropy function, and $\mathds{1}_{Z_i}$ stands for the unit vector for the $Z_i$-th coordinate in $\mathbb{R}^{|\mcZ|}$. Hence, for each data index $i$, $\mathbf{L}_{\text{new}}^\top\mathds{1}_{Z_i}\in\mathbb{R}_+^{|\mcS|}$, which is a \emph{random} vector, is treated as the target ``pseudo-label'' vector for the input (context) $\Cb_i$. Note the pseudo-label is not a unit vector as in the case of the usual supervised multi-class classification. 
For learning the parameter $\wb$,
the ordinary back-propagation and variants of mini-batch SGD are used to minimize the objective function. Since the pseudo-label $\Ellb_{\text{new}}(Z,s)$ is negatively correlated with $\Lb(x,s(Z))$ in expectation by design (following from (\ref{eq:L_max}) and the unbiased property of $\Ellb$), the network will tend to assign higher probability for the mapping $s$ that has high $\Ellb_{\text{new}}(Z_i,s)$ value for each $\Cb_i$. 

Once (\ref{eq:objective}) converges after sufficient number of iterations, the converged parameter is denoted as $\tilde{\wb}$. Then, the single-letter mapping defined by Neural DUDE for the context $\Cb\in\Cb[k]$ is expressed as 
$
s_{k,\text{N-DUDE}}(\Cb,\cdot)=\arg\max_{s\in\mcS}\mathbf{p}^k(\tilde{\wb},\Cb)_s,
$
and the reconstruction at location $i$ becomes $\hat{X}_{i,\text{N-DUDE}}(Z^n)=s_{k,\text{N-DUDE}}(\Cb_i,Z_i)$. Hence, in summary, Neural DUDE denoises the noisy data after adaptively training the network parameters with the \emph{same} noisy data, without requiring any additional supervised training set. \citep{MooMinLeeYoo16} shows encouraing empirical results including the robustness of the performance with respect to $k$, and in this paper, we provide theoretical justification of Neural DUDE.

%% file: main_arxiv.tex
\section{Main Results}\label{sec:main results}

Before stating our main theorem, we introduce additional notations. First, 
denote $\mathcal{W}_k^{L,N}$ as the set of the parameters of the neural network that has the architecture in Figure \ref{fig:n_dude_arch}, namely, takes $\Cb\in\Cb[k]$ as input and has $L$ layers and $N$ nodes. Note $N=\sum_{\ell=1}^Ln_\ell$. By denoting $\wb_{\ell,m}\in\mathbb{R}^{n_{\ell-1}}$ as the weight parameter vector associated with the $m$-th node in the $\ell$-th layer in $\mathcal{W}_k^{L,N}$, we assume $\|\wb_{\ell,m}\|_2\leq B$ with $B<\infty$ for all $\ell$ and $m$.


For each $\wb\in\mathcal{W}_k^{L,N}$ and the context $\Cb\in\Cb[k]$, we denote
\be
s_k[\mathbf{w}](\Cb,\cdot)=\arg\max_{s\in\mathcal{S}} \mathbf{p}^k(\mathbf{w},\Cb)_s\label{eq:s_tilde_def_2}
\ee 
as the single-symbol denoiser that is defined by the context $\Cb\in\Cb[k]$, the network parameters $\mathbf{w}$, and the neural network architecture $\mathbf{p}^k(\mathbf{w},\cdot)$. 
Note the notation in (\ref{eq:s_tilde_def_2}) highlights the dependency on $\wb$. 
Also, for brevity, we denote
\be
\bar{\Ellb}_n[\wb] &\triangleq& \frac{1}{n}\sum_{i=1}^n\Ellb(Z_i,s_k[\wb](\Cb_i,\cdot))\label{eq:avg_est_loss}\\
\bar{\Lb}_n[\wb] &\triangleq& \frac{1}{n}\sum_{i=1}^n\Lb(x_i,s_k[\wb](\Cb_i,Z_i))\label{eq:ave_true_loss}
\ee
as the average estimated loss and average true denoising loss for the sliding window denoiser $s_k[\mathbf{w}](\cdot)$, respectively. Now, we have the following main theorem of the paper. 

\begin{theorem}\label{eq:main_thm}
Let $\hat{\wb}=\arg\min_{\wb\in\mathcal{W}_{k}^{L,N}} \bar{\Ellb}_n[\wb]$ 
 and $\wb^{\star}=\arg\min_{\wb\in\mathcal{W}_{k}^{L,N}} \bar{\Lb}_n[\wb]$. Also, let $\delta>0$. Then, for all $x^n\in\mcX^n$, with probability at least 1-$\delta$, we have
\begin{align}
&\bar{\Lb}_n[\hat{\wb}]-\bar{\Lb}_n[\wb^\star]\nonumber\\
\leq&\ 2C_{\max}\Bigg(4|\mcS|\sqrt{\tilde{C}\sqrt{\frac{k}{n}}}+(2k+1)\sqrt{\frac{2\log (2/\delta)}{n}}\Bigg)\label{eq:thm_bound}
\end{align}
in which $C_{\max}\triangleq \max_{z,s}|\Ellb(z,s)|+\max_{x,\hat{x}}|\Lb(x,\hat{x})|$ and $\tilde{C}=(2B)^{L+1}\sqrt{\big(\prod_{\ell=1}^{L}n_\ell\big)\frac{|\mcS|}{2}}$.
\end{theorem}
\emph{Remark:} The theorem states that the average denoising loss of $\hat{\wb}$, which is the minimizer of the average \emph{estimated} loss in $\mathcal{W}_{k}^{L,N}$, can be upper bounded by the best possible denoising loss in $\mathcal{W}_{k}^{L,N}$, $\bar{\Lb}_n[\wb^\star]$ (bias, or the approximation error), plus the right-hand side of the inequality in (\ref{eq:thm_bound}) (variance, or the estimation error), with high probability. Therefore, we can interpret the theorem similarly as the well-known bias-variance tradeoff in supervised learning; namely, as the neural network architecture becomes more complex, the bias term will decrease, but the varaince term will increase.  Hence, in order to achieve small $\bar{\Lb}_n[\hat{\wb}]$, we need to control the model complexity of $\mathcal{W}_k^{L,N}$ to optimize the bias-variance tradeoff. As mentioned in the Introduction, the difference between supervised learning and our problem is that while the supervised learning cares about the prediction performance for the unseen test data, we care about the denoising performance for the \emph{unseen underlying clean $x^n$}. 
Moreover, although $\hat{\wb}$ is not practically attainable due to the non-convex objective of (\ref{eq:avg_est_loss}), above theorem justifies the performance of Neural DUDE as follows; the objective function (\ref{eq:objective}) that Neural DUDE uses for training is a convex surrogate of (\ref{eq:avg_est_loss}) that ensures the infinite-sample consistency (ISC) \citep[Section 4.4.3]{Zhang04b} when the neural network has a single layer. Namely, when $n\rightarrow \infty$, $\bar{\Ellb}_n[\tilde{\wb}]\rightarrow\bar{\Ellb}_n[\hat{\wb}]$ in probability for a single-layer network. Thus, we can expect that the $\wb$ that tries to minimize (\ref{eq:objective}) will also minimize (\ref{eq:avg_est_loss}), which parallels the usage of cross-entropy as an objective function to minimize the multi-class classification error in supervised learning. 

The key tool for proving Theorem \ref{eq:main_thm} is the following uniform concentration result. 
\begin{theorem}\label{thm:uniform_conv}
Let $\delta, \gamma>0$ and consider $C_{\max}$ and $\tilde{C}$ defined in Theorem \ref{eq:main_thm}. Then, for all $x^n\in\mcX^n$ , with probability at least 1-$\delta$,
\begin{align}
&\sup_{\wb\in\mathcal{W}_k^{L,N}} \big|\bar{\Ellb}_n[\wb]-\bar{\Lb}_n[\wb] \big|\nonumber\\
\leq& \ \ C_{\max} \Bigg(2\gamma|\mcS|^2+\frac{2\tilde{C}}{\gamma}\sqrt{\frac{k}{n}}+ (2k+1)\sqrt{\frac{2\log(2/\delta)}{n}}\Bigg).\nonumber
\end{align}
\end{theorem}
\emph{Remark:} Note $\gamma$ in the theorem is a free parameter, and the bound in Theorem \ref{eq:main_thm} is obtained by optimizing $\gamma$ in the above bound. Furthermore, for fixed constants, we observe that $\bar{\Ellb}_n[\wb]$ concentrates on $\bar{\Lb}_n[\wb]$ \emph{uniformly} for \emph{all} $\wb\in\mathcal{W}_k^{L,N}$ and $x^n\in\mcX^n$, provided that $k=o(\sqrt{n})$. Note a similar concentration result obtained by the information-theoretic method in \citep[Examples 5 and 6]{Ord13} cannot be applied for Neural DUDE, since the denoised symbol in Neural DUDE is affected by the entire noisy data through the training process. Above theorem supports the experimental finding in Figure \ref{fig:dude_est_ber} and \ref{fig:n_dude_est_ber}; namely, the concentration of the average estimated loss on the average denoising loss for Neural DUDE happens for much larger $k$ than for DUDE, when the model architecture is simple and the alphabet size is small. 

For comparison, following proposition, of which proof is given in the Supplementary Material, shows the weak concentration property of DUDE in \citep{weissman2005universal}.



\begin{proposition}\label{prop:dude}
Let $\mcS_k$ denote the class of all the $k$-th order sliding window denoisers, $s_k$, and let $\delta>0$. Then, for all $x^n\in\mcX^n$, with probability at least $1-\delta$, we have
\begin{align}
\max_{s_k\in\mcS_k}&\frac{1}{n-2k}\sum_{i=k+1}^{n-k}[\Ellb(Z_i,s_k(\Cb_i,\cdot))-\Lb(x_i,s_k(\Cb_i,Z_i))]\nonumber\\
=&\ O\left(\sqrt{\frac{k|\mcZ|^{2k}\log(|\mcS|/\delta)}{n}}\right).\label{eq:big-o}
\end{align}
\end{proposition}
\emph{Remark:} We observe that (\ref{eq:big-o}) vanishes when $k=o(\log n)$. Since DUDE with window size $k$ is also in $\mcS_k$, the proposition gives the justification of the poor concentration property of DUDE with respect to $k$ for fixed $n$ as in Figure \ref{fig:dude_est_ber}. 


%% file: proof_arxiv.tex

\section{Proof of the Main Results}\label{sec:proof}



\begin{figure*}[h]
    \centering
    \subfigure[A probability simplex $\Delta^{|\mcS|}$.]{\label{fig:simplex}
    \includegraphics[width=0.3\textwidth]{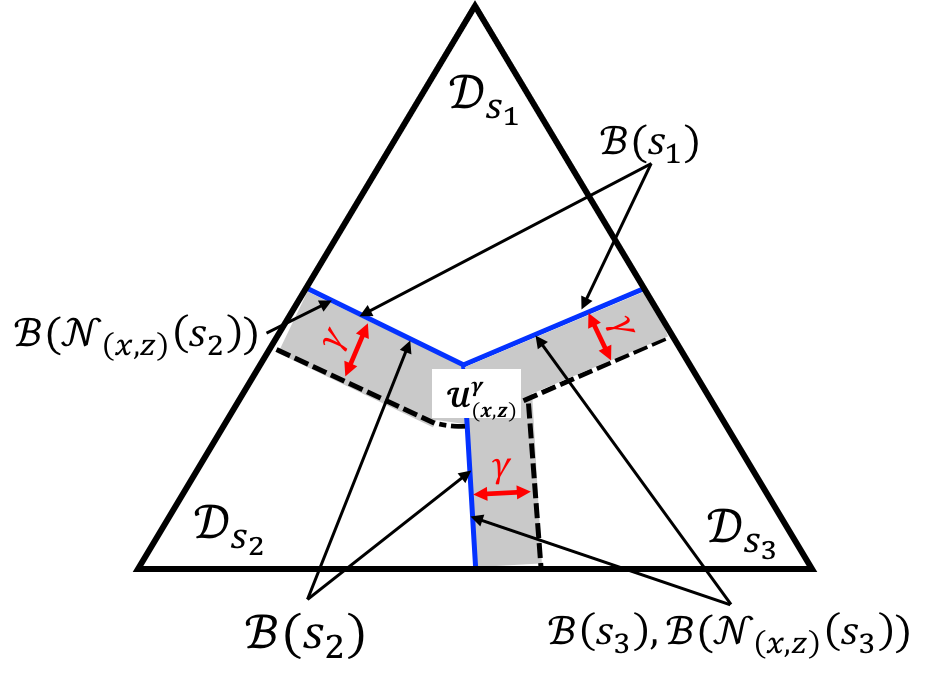}}\hspace{.4in}
    \subfigure[$\tilde{r}^\gamma_{(x,z)}(\mathbf{p})$ on $\mathbf{p}\in\Delta^{|\mcS|}$.]{\label{fig:lipschitz}
    \includegraphics[width=0.3\textwidth]{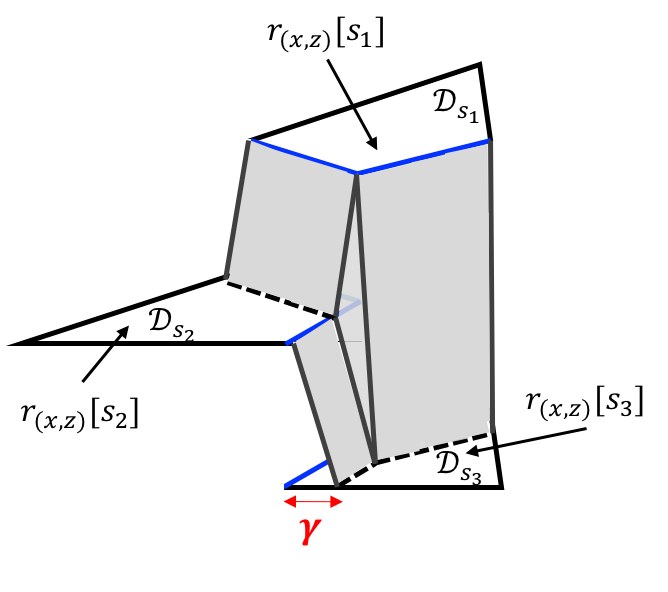}}
    \caption{An example for $|\mcS|=3$. We assume $r_{(x,z)}[s_1]\geq r_{(x,z)}[s_2] \geq r_{(x,z)}[s_3]$, in which we have $\mathcal{N}_{(x,z)}(s_1)=\phi, \mathcal{N}_{(x,z)}(s_1)=\{s_1\}, \mathcal{N}_{(x,z)}(s_1)=\{s_1,s_2\}$. (a) The $\gamma$-margin $\mathcal{U}_{(x,z)}^\gamma$ in (\ref{eq:gamma_margin}) is shown as the shaded region, and the decision region and boundary for each region are also shown. (b) The interpolated regret function $\tilde{r}_{(x,z)}^\gamma(\mathbf{p})$ in (\ref{eq:regret_fcn_2}), which is Lipschitz continuous in $\mathbf{p}$ on $\Delta^{|\mcS|}$, is shown.}
    \label{fig:simplex1}
\end{figure*}

The main gist of the proving the theorems consists of three parts. First, since $\sup_{\wb\in\mathcal{W}_k^{L,N}} \big|\bar{\Ellb}_n[\wb]-\bar{\Lb}_n[\wb] \big|$ is not a continuous function in $\wb$, we consider a Lipschitz continuous upper bound of it and show the upper bound concentrates on its expectation (Lemma \ref{lem:sup_concent}). Second, we bound the expectation considered in Lemma \ref{lem:sup_concent} basically with the Rademacher complexity of the neural networks in $\mathcal{W}_{k}^{L,N}$ (Lemma \ref{lem:rademacher}). Third, we obtain the upper bound on the Rademacher complexity (Lemma \ref{lem:rc_bound}). Using the results of the three lemmas, we prove Theorem \ref{thm:uniform_conv} followed by Theorem \ref{eq:main_thm}. Now, we introduce a few more notations and definitions necessary for stating and proving the lemmas.

\subsection{Additional notations and definitions}\label{subsec:add_notation}

For any $x\in\mcX$, $z\in\mcZ$ and $s\in\mcS$, we define the \emph{per-symbol regret} as
$
r_{(x,z)}[s] \triangleq \Ellb(z,s)-\Lb(x,s(z)).
$
Moreover, for schemes that determine the single-symbol denoiser by finding the maximum argument of a probability vector $\mathbf{p}\in\Delta^{|\mcS|}$ as in Neural DUDE, we also denote 
$
r_{(x,z)}(\mathbf{p}) \triangleq r_{(x,z)}[\mcS(\mathbf{p})]
$as the \emph{regret function} (in $\mathbf{p}$), by defining $\mcS(\mathbf{p})=s$ if $$\mathbf{p}\in\mathcal{D}_s\triangleq \{\mathbf{p}\in\Delta^{|\mcS|}: s=\arg\max_{s'\in\mcS}\mathbf{p}_{s'}\}.$$ Namely, $\mathcal{D}_s$ is the decision region for $s$.

With above notations and (\ref{eq:s_tilde_def_2}), we can express the $i$-th per-symbol regret as 
\be
&&\mathbf{L}(Z_i,s_k[\mathbf{w}](\Cb_i,\cdot))-\Lb(x_i,s_k[\mathbf{w}](\Cb_i,Z_i)\nonumber\\
&=&r_{(x_i,Z_i)}(\mathbf{p}^k(\wb,\Cb_i))\triangleq r_i[\wb] \label{eq:new_regert}.
\ee
In (\ref{eq:new_regert}), we introduced the notation $r_i[\wb]$ for brevity and to highlight the dependency on $\wb$. Now, by denoting 
$$
\mathbf{R}_n[\wb]\triangleq \bar{\Ellb}_n[\wb]-\bar{\Lb}_n[\wb] = \frac{1}{n}\sum_{i=1}^nr_i[\wb],
$$
it becomes clear that Theorem \ref{thm:uniform_conv} obtains a bound on 
$
\sup_{\wb\in\mathcal{W}_{k}^{L,N}}|\mathbf{R}_n[\wb]|.
$

One important point is that the regret function $r_{(x,z)}(\mathbf{p})$ defined above is not a continuous function in $\mathbf{p}$ for any $x$ and $z$. Due to the technical necessity required in the later analyses, we need to define another function $\tilde{r}_{(x,z)}^\gamma(\mathbf{p})$, which is a pointwise upper bound on $r_{(x,z)}(\mathbf{p})$ and is Lipschitz continuous in $\mathbf{p}$. To do that, for each $s\in\mcS$, we first let
\be
\mathcal{B}(s) \triangleq \{\mathbf{p}\in\Delta^{|\mcS|}:\mathbf{p}_s=\max_{s'\neq s}\mathbf{p}_{s'}\}\nonumber
\ee
be the set of the decision boundaries for $\mathcal{D}_s$ defined above.
Furthermore, for each clean-noisy pair $(x,z)$ and a single-symbol denoiser $s$, define
\be
\mathcal{N}_{(x,z)}(s)&\triangleq&\{s'\in\mcS:\mathcal{B}(s')\cap\mathcal{B}(s)\neq \phi\nonumber\\
&&\ \ \ \text{and}\ r_{(x,z)}[s']\geq r_{(x,z)}[s]\}\nonumber
\ee
as the set of the neighboring single symbol denoisers that share decision boundaries with $s$ and has larger per-symbol regret for $(x,z)$. Then, we define 
\be
\mathcal{B}(\mathcal{N}_{(x,z)}(s)) \triangleq \bigcup_{s'\in \mathcal{N}_{(x,z)}(s)}\mathcal{B}(s') \cap \mathcal{B}(s)\label{eq:subset_boundaries}
\ee
as a subset of $\mathcal{B}(s)$ that only contains boundaries between $s$ and the single-symbol denoisers in $\mathcal{N}_{(x,z)}(s)$. Note when $\mathcal{N}_{(x,z)}(s)=\phi$, that is, when $s=\arg\max_{s'\in\mcS}r_{(x,z)}[s']$, then $\mathcal{B}(\mathcal{N}_{(x,z)}(s))=\phi$ as well. Now, for small $\gamma>0$, we define the \emph{$\gamma$-margin} for $(x,z)$ as
\be
\mathcal{U}_{(x,z)}^\gamma\triangleq \{\mathbf{p}\in\Delta^{|\mcS|}: \textbf{dist}\Big(\mathbf{p},\mathcal{B}(\mathcal{N}_{(x,z)}(\mcS(\mathbf{p}))\Big) \leq \gamma \},\label{eq:gamma_margin}
\ee
in which 
$
\textbf{dist}(\mathbf{p},\mathcal{E})\triangleq \min_{\mathbf{p}'\in\mathcal{E}}\|\mathbf{p}-\mathbf{p}'\|_2
$
for a set $\mathcal{E}\in\Delta^{|\mcS|}$. Also, we define $\text{dist}(\mathbf{p},\phi)=\infty$. In words, $\mathcal{U}_{(x,z)}^\gamma$ is the set of the probability vectors that are within distance $\gamma$ from the boundaries for the single-symbol denoisers that have larger per-symbol regrets than $\mcS(\mathbf{p})$. A simple example that describes the defined notations for $|\mcS|=3$ is given in Figure \ref{fig:simplex}.

With the definition (\ref{eq:gamma_margin}), we now define the \emph{interpolated} regret function for $\mathbf{p}\in\Delta^{|\mcS|}$
\be
\tilde{r}_{(x,z)}^\gamma(\mathbf{p}) \triangleq  r_{(x,z)}(\mathbf{p})+\Delta r_{(x,z)}^\gamma(\mathbf{p}),\label{eq:regret_fcn_2}
\ee
in which 
$\Delta r_{(x,z)}^\gamma(\mathbf{p})\geq 0$ for all $\mathbf{p}\in\Delta^{|\mcS|}$ and is defined as
\be
&&\Delta r_{(x,z)}^\gamma(\mathbf{p}) \label{eq:delta_r}\\
&\triangleq&\begin{cases}
\textbf{L.I.} \Big(r_{(x,z)}[\mcS(\mathbf{p})], \mathcal{N}_{(x,z)}(\mcS(\mathbf{p})) \Big) &\ \text{if} \ \mathbf{p}\in\mathcal{U}_{(x,z)}^{\gamma}\\
0 & \ \text{otherwise}.
\end{cases}\nonumber
\ee
The notation $\textbf{L.I.} \big(r_{(x,z)}[\mcS(\mathbf{p})], \mathcal{N}_{(x,z)}(\mcS(\mathbf{p})) \big)$ in (\ref{eq:delta_r}) stands for the appropriate linear interpolation value between the per-symbol regret value $r_{(x,z)}[\mcS(\mathbf{p})]$ and the larger values in $\{r_{(x,z)}[s]:s\in \mathcal{N}_{(x,z)}(\mcS(\mathbf{p}))\}$, determined by $\textbf{dist}(\mathbf{p},\mathcal{B}(\mathcal{N}_{(x,z)}(\mcS(\mathbf{p})))$. While the specific function form in (\ref{eq:delta_r}) is not important, the important part is that $\tilde{r}_{(x,z)}^\gamma(\mathbf{p}) $ becomes Lipschitz continuous in $\mathbf{p}$; that is, for the constant $C_{\max}\triangleq \max_{z,s}|\Ellb(z,s)|+\max_{x,\hat{x}}|\Lb(x,\hat{x})|$, we have
\be
|\tilde{r}_{(x,z)}^\gamma(\mathbf{p})-\tilde{r}_{(x,z)}^\gamma(\mathbf{p}')|\leq \frac{C_{\max}}{\gamma}\|\mathbf{p}-\mathbf{p}'\|_2\label{eq:lipschitz_property}
\ee
for all $\mathbf{p},\mathbf{p}'\in\Delta^{|\mcS|}$.
The property (\ref{eq:lipschitz_property}) becomes necessary in proving Lemma \ref{lem:rc_bound} below and Theorem \ref{thm:uniform_conv}. The example of $\tilde{r}_{(x,z)}^\gamma(\mathbf{p})$ on $\Delta^{|\mcS|}$ is shown in Figure \ref{fig:lipschitz}.

\subsection{Three lemmas}\label{subsec:lemmas}

With (\ref{eq:regret_fcn_2}), define the $i$-th interpolated per-symbol regret as 
\be
\tilde{r}^\gamma_i[\wb]\triangleq\tilde{r}^\gamma_{(x_i,Z_i)}(\mathbf{p}^k(\wb,\Cb_i))\label{eq:r_tilde}
\ee
and the corresponding Lipschitz-continuous average regret as $\tilde{\mathbf{R}}^\gamma_n[\wb]=\frac{1}{n} \sum_{i=1}^n\tilde{r}^\gamma_i[\wb]$. Since $\tilde{r}^\gamma_{(x,z)}(\mathbf{p})\geq r_{(x,z)}(\mathbf{p})$ for all $(x,z)$ and $\mathbf{p}$, we have $\tilde{\mathbf{R}}^\gamma_n[\wb]\geq \mathbf{R}_n[\wb]$ \emph{a.s.} for all $\wb\in\mathcal{W}^{L,N}_k$.
Thus, we have the upper bound
\be
\sup_{\wb\in\mathcal{W}_k^{L,N}} \mathbf{R}_n[\wb] \leq \sup_{\wb\in\mathcal{W}_k^{L,N}} \tilde{\mathbf{R}}^\gamma_n[\wb] \triangleq G_n\ \ \ \ a.s.\label{eq:pointwise_ub}
\ee
Note the bound in (\ref{eq:pointwise_ub}) is in the almost sure sense since both $\mathbf{R}_n[\wb]$ and $\tilde{\mathbf{R}}^\gamma_n[\wb]$ are random variables. In (\ref{eq:pointwise_ub}), we defined $G_n$ to denote the right-hand side of the inequality. 
Now, before proving our theorem, we present three lemmas which are essentially on bounding $G_n$.
The full proofs of the lemmas are given in the Supplementary Material, and we only give the high-level proof sketches below.



\begin{lemma}\label{lem:sup_concent}
For $\epsilon>0$ and $C_{\max}\triangleq \max_{z,s}|\Ellb(z,s)|+\max_{x,\hat{x}}|\Lb(x,\hat{x})|$, we have
\be
\mathbb{P}\Big(G_n-\mathbb{E}(G_n)>\epsilon \Big)\leq \exp\Big(-\frac{n\epsilon^2}{2(2k+1)^2C_{\max}^2}\Big).
\ee
\end{lemma}
\emph{Proof sketch:} The proof follows from applying the McDiarmid's inequality \citep{McDiarmid89} and the independence of $(Z_1,\ldots,Z_n)$ given the individual clean sequence $x^n$. \qed

\begin{lemma}\label{lem:rademacher}
Let $\cb=(z_{-k}^{-1},z_1^k)\in\Cb[k]$ and let $\mathcal{A}\triangleq\{(x,z_0,\cb)\rightarrow \tilde{r}^\gamma_{(x,z_0)}(\mathbf{p}^k(\wb,\cb)): \mathbf{w}\in\mathcal{W}_k^{L,N}\}$ be the function class of the interpolated per-symbol regret functions, parameterized by $\mathbf{w}\in\mathcal{W}_k^{L,N}$.
Then, for $\gamma>0$ used in (\ref{eq:gamma_margin}) and $C_{\max}$ defined in Lemma \ref{lem:sup_concent}, we have
\be
\mathbb{E}(G_n)\leq 2\big(\mathcal{R}_n(\mathcal{A})+C_{\max}\gamma|\mcS|^2\big),\label{eq:ineq_lem2}
\ee
in which $\mathcal{R}_n(\mathcal{A})$ is the Rademacher complexity of $\mathcal{A}$,
\be
\mathcal{R}_n(\mathcal{A})\triangleq\mathbb{E}\Big(\sup_{\mathbf{w}\in\mathcal{W}_k^{L,N}}\frac{1}{n}\sum_{j=1}^n\sigma_i\tilde{r}^\gamma_i[\wb]\Big).\label{eq:rc_definition}
\ee
 In (\ref{eq:rc_definition}), $\{\sigma_i\}_{i=1}^n$ are drawn i.i.d. uniform over $\{+1,-1\}$, and $\tilde{r}^\gamma_i[\wb]$ is as defined in (\ref{eq:r_tilde}).
 \end{lemma}
\emph{Proof sketch:} We utilize the fact that $\mathbb{E}(r_i[\wb])=0$ for all $i$, which follows from the unbiased property of $\Ellb(Z,s)$. Furthermore, by bounding $\Delta r_i^{\gamma}[\wb]$ and following the standard symmetrization argument, we obtain the upper bound (\ref{eq:ineq_lem2}).\qed

\begin{lemma}\label{lem:rc_bound}
For $\gamma>0$ used in (\ref{eq:gamma_margin}), we have the following bound on $\mathcal{R}_n(\mathcal{A})$ in (\ref{eq:rc_definition}):
\be
\mathcal{R}_n(\mathcal{A}) \leq \frac{C_{\max}\tilde{C}}{\gamma}\sqrt{\frac{k}{n}}.\label{eq:lem3_bound}
\ee
in which $\tilde{C}=(2B)^{L+1}\sqrt{\big(\prod_{\ell=1}^{L}n_\ell\big)\frac{|\mcS|}{2}}$.
\end{lemma}
\emph{Proof sketch:} We iteratively apply the Lipschitz composition property \citep[Corollary 3.17]{LedTal91} for Rademacher complexity and utilize the property of the used network architecture. \ \qed

\subsection{Proof of the theorems}
We now use above lemmas and prove the main theorems. \\
\emph{Proof of Theorem \ref{thm:uniform_conv}:} First, we have
\begin{align}
\mathbb{P}\Big(\sup_{\wb\in\mathcal{W}_k^{L,N}} \big|\mathbf{R}_n[\wb]& \big|\geq \epsilon\Big)
\leq\mathbb{P}\Big(\sup_{\wb\in\mathcal{W}_k^{L,N}}\mathbf{R}_n[\wb]\geq \epsilon\Big)\nonumber\\
&+\mathbb{P}\Big(\sup_{\wb\in\mathcal{W}_k^{L,N}}(-\mathbf{R}_n[\wb])\geq \epsilon\Big)\label{eq:bounds}
\end{align}
by applying the union bound. Now, from the pointwise upper bound property in (\ref{eq:pointwise_ub}), we have
\be
(\text{The first term in (\ref{eq:bounds})})
&\leq&\mathbb{P}\Big(\sup_{\wb\in\mathcal{W}_k^{L,N}}\tilde{\mathbf{R}}^\gamma_n[\wb]\geq \epsilon\Big)\nonumber\\& = &\mathbb{P}(G_n\geq \epsilon),
\ee
in which we used the definition of $G_n$ given in (\ref{eq:pointwise_ub}). By combining the results of Lemma \ref{lem:sup_concent} and Lemma \ref{lem:rademacher}, we obtain
\begin{align}
&\mathbb{P}(G_n\geq \epsilon)
\leq \exp\Bigg(-\frac{n\big(\epsilon-\mathbb{E}(G_n)\big)^2}{2(2k+1)^2C_{\max}^2}   \Bigg)\nonumber\\
\leq&\exp\Bigg(-\frac{n\big(\epsilon-(2\mathcal{R}_n(\mathcal{A})+2C_{\max}\gamma|\mcS|^2)\big)^2}{2(2k+1)^2C_{\max}^2}   \Bigg)\triangleq \frac{\delta}{2}\label{eq:def_delta}
\end{align}
where the inequality (\ref{eq:def_delta}) holds for $\epsilon\geq 2\mathcal{R}_n(\mathcal{A})+2C_{\max}\gamma|\mcS|^2$.

Now, for the second term in (\ref{eq:bounds}), we can analogously define a function, $\hat{r}^\gamma_i[\wb]$, that upper bounds the negated per-symbol regret function, $-r_i[\wb]$, as in (\ref{eq:regret_fcn_2}) and (\ref{eq:r_tilde}). Then, we obtain the indetical result as Lemma \ref{lem:rademacher} for the function class of interpolated per-symbol negated regret functions. That is, with defining $\hat{G}_n\triangleq\sup_{\wb\in\mathcal{W}}(\frac{1}{n}\sum_{i=1}^n\hat{r}^\gamma_i[\wb])$, and $\hat{\mathcal{A}}\triangleq\{(x,z_0,\cb)\rightarrow \hat{r}^\gamma_{(x,z_0)}(\mathbf{p}^k(\wb,\cb)): \wb\in\mathcal{W}_k^{L,N}\}$ as the function class for the negated regret functions, we obtain
$
\mathbb{E}(\hat{G}_n)\leq 2\mathcal{R}_n(\hat{\mathcal{A}})+2C_{\max}\gamma|\mcS|^2,
$
via the same derivation in Lemma \ref{lem:rademacher}. Since it is clear that 
$\mathcal{R}_n(\hat{\mathcal{A}})$ and $\mathcal{R}_n(\mathcal{A})$ have the same upper bound, we obtain
$
(\text{The second term in (\ref{eq:bounds})})\leq  \mathbb{P}(\hat{G}_n\geq \epsilon)\leq \frac{\delta}{2},
$
in which $\delta$ is defined in (\ref{eq:def_delta}).

Now, by solving for $\epsilon$, we obtain
\begin{align}
\epsilon=&2\mathcal{R}_n(\mathcal{A})+2C_{\max}\gamma|\mcS|^2 +(2k+1)C_{\max}\sqrt{\frac{2\log(2/\delta)}{n}}\nonumber,
\end{align}
and by plugging in the result of Lemma \ref{lem:rc_bound}, we have proved the theorem. \qed


\emph{Proof of Theorem \ref{eq:main_thm}:} From the definitions of $\hat{\wb}$ and $\wb^\star$, we have
\begin{align}
&\bar{\Lb}_n[\hat{\wb}]-\bar{\Lb}_n[\wb^\star]\nonumber\\
=& \bar{\Lb}_n[\hat{\wb}]-\bar{\Ellb}_n[\hat{\wb}]+\bar{\Ellb}_n[\hat{\wb}]-\bar{\Ellb}_n[\wb^\star]+\bar{\Ellb}_n[\wb^\star]-\bar{\Lb}_n[\wb^\star]\nonumber\\
\leq& \big(\bar{\Lb}_n[\hat{\wb}]-\bar{\Ellb}_n[\hat{\wb}]\big)+\big(\bar{\Ellb}_n[\wb^\star]-\bar{\Lb}_n[\wb^\star]\big)\label{eq:thm1_proof_1}\\
\leq& 2\ \bigg(\sup_{\wb\in\mathcal{W}_k^{L,N}} \big|\bar{\Ellb}_n[\wb]-\bar{\Lb}_n[\wb] \big|\bigg)\label{eq:thm1_proof_2}
\end{align}
almost surely, in which the first equality follows from subtracting and adding the same terms, (\ref{eq:thm1_proof_1}) follows from $\bar{\Ellb}_n[\hat{\wb}]\leq\bar{\Ellb}_n[\wb^\star]$ by definition, and (\ref{eq:thm1_proof_2}) follows from taking supremum over $\mathcal{W}_k^{L,N}$. Therefore, we have
\be
\mathbb{P}\Big(\bar{\Lb}_n[\hat{\wb}]-\bar{\Lb}_n[\wb^\star]\geq\epsilon\Big)\nonumber\leq\mathbb{P}\bigg(\sup_{\wb\in\mathcal{W}_k^{L,N}} \big|\mathbf{R}_n[\wb] \big|\geq \frac{\epsilon}{2}\bigg)\nonumber,
\ee
and using the result of Theorem \ref{thm:uniform_conv} together with
\be
\gamma|\mcS|^2+\frac{\tilde{C}}{\gamma}\sqrt{\frac{k}{n}}\geq 2|\mcS|\sqrt{\tilde{C}\sqrt{\frac{k}{n}}}\nonumber
\ee 
for all $\gamma>0$, we obtain the bound in Theorem \ref{eq:main_thm}.\ \qed

\emph{Remark:} We note that the constant $\tilde{C}$ can be quite large when the number of layers, $L$, and the number of nodes $N$ of the neural network grows. Such dependency is the artifact from bounding the Rademachaer complexity $\mathcal{R}(\mathcal{A})$, and an improved bound such as \citep{AndPanValZha14} can be also used. The point of the Theorem \ref{eq:main_thm} is to show the existence of the bias-variance tradeoff for the denoising loss of the neural network based denoisers learned by minimizing the average estimated loss. Furthermore, through the experiments in the next section, we show the dependency of $\tilde{C}$ on $L$ and $N$ is not as severe as obtained in the theorem. 

%% file: experiments_arxiv.tex
\section{Experiments}\label{sec:experiments}
\begin{figure*}[h]
    \centering
    
    \subfigure[True\&estimated BER/$\delta$ for test set]{\label{fig:varying_epoch}
    \includegraphics[width=0.31\textwidth]{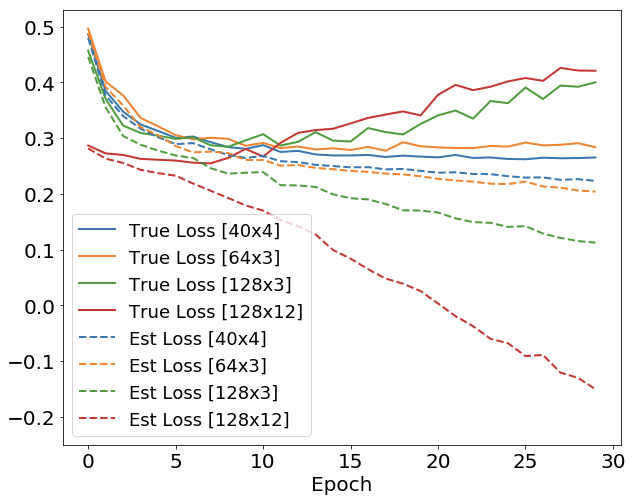}}
    \subfigure[Objective function (\ref{eq:objective}) for test set]{\label{fig:training_error}
    \includegraphics[width=0.31\textwidth]{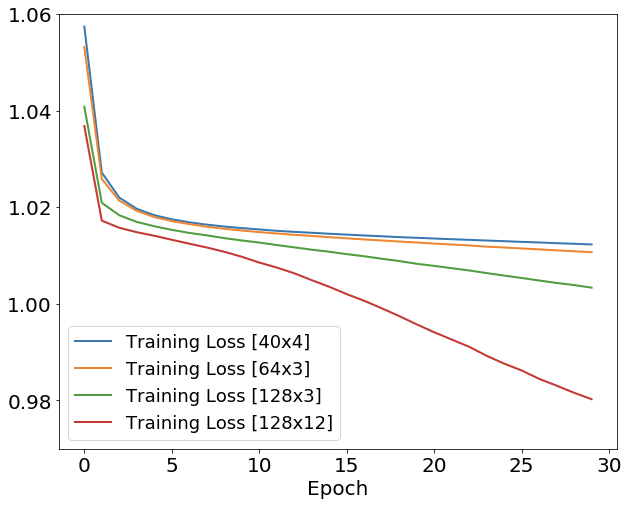}}
    \subfigure[True\&estimated BER/$\delta$ for valid. set]{\label{fig:varying_k}
    \includegraphics[width=0.31\textwidth]{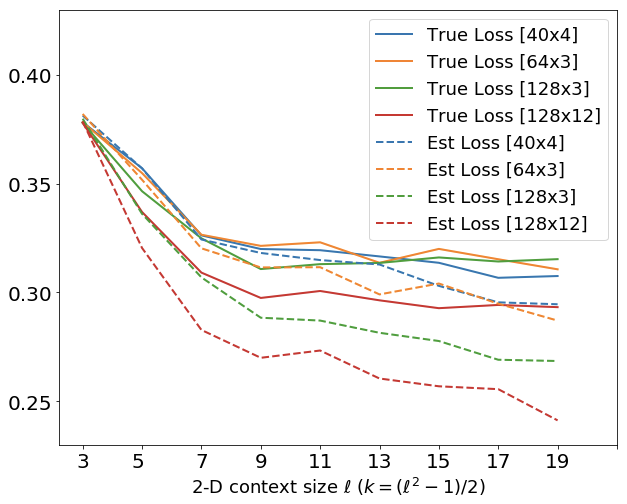}}
    \caption{(a) True and estimated BER/$\delta$ for varying epochs and model architecture on the test set. $\ell=11$ was fixed. (b) The objective function (\ref{eq:objective}) for the test set with $\ell=11$.(c) True and estimated BER/$\delta$ for varying $\ell$ on the validation set. The epoch was fixed to $7$. }
    \label{fig:experiment_result}
\end{figure*}

In this section, we carry out experiments regarding binary image denoising to corroborate the theoretical results of this paper. As in \citep[Section 5.2]{MooMinLeeYoo16}, we tested the denoising performance on the five images with various textual characteristics and sizes, \ie, Einstein, Lena, Barbara, Cameraman, Shannon\footnote{The images are shown in the Supplementary Material.}. Einstein  is a halftone image with size 256$\times$256, Shannon is a scanned text image with size 256$\times$256, and the rest three  are binarized natural benchmark images with 512$\times$512. To simplify the experiment, we assumed $\Pib$ is the binary symmetric channel (BSC) with crossover probability $\delta=0.1$ and used the Hamming loss matrix as $\Lb$. Hence, the average denoising loss in this case becomes the Bit Error Rate (BER). All the numbers we report below is the relative BER compared to $\delta$. The optimizer for the neural network was fixed to Adam \citep{KinBa15} with default learning rate $10^{-3}$. While the experimental setting is simple, it is enough to show the general phenomenon outlined by the result of this paper. \

The original experiments in \citep{MooMinLeeYoo16} considered the raster scanned 1-D data of the images and varied the size of the 1-D context ($k$) while fixing the model architecture and the number of epochs for minimizing (\ref{eq:objective}). In this paper, we considered the two dimensional (2-D) context $\Cb_i^{\ell\times\ell}$, an $\ell\times\ell$ square patch around the noisy symbol $Z_i$ that does \emph{not} include $Z_i$, since it can more naturally capture the contextual information from images. Also, the boundaries were zero-padded such that the denoising can be done at all locations. Note when $\ell$ is odd, 1-D context with $k=(\ell^2-1)/2$  has the same data size as $\Cb_i^{\ell\times\ell}$. Thus, we can compare the $k$ values in \citep[Table 1]{MooMinLeeYoo16} with our results. 

Since Theorem \ref{eq:main_thm} shows that bias-variance tradeoff exists for the denoising loss of Neural DUDE, we checked the performance with more varying hyperparameters in addition to $k$ (or $\ell$): the number of layers, the number of nodes in each layer, and the number of epochs. For model architectures, we tried following four models; 4 layers with 40 nodes in each layer, 3 layers with 64 nodes, 3 layers with 128 nodes, and 12 layers with 128 nodes. 
Above models are dubbed as $40\times 4$, $64\times 3$, $128\times 3$, and $128\times 12$, respectively. For the number of epochs, we tried until 30 epochs, and for the 2-D context size, we varied $\ell=\{3,5,7,9,11,13,15,17,19\}$, which translates to $k=\{4,12,24,40,60,84,112,144,180\}$ for 1-D context.

Figure \ref{fig:varying_epoch} shows the true ($\bar{\Lb}[\wb]$) and estimated BER ($\bar{\Ellb}[\wb]$) averaged over the 5 test images for each tested model, with respect to fixed $\ell=11$ and varying epochs. Furthermore, Figure \ref{fig:training_error} shows the average objective function values (\ref{eq:objective}) for the same setting. Firstly, by comparing both figures, we can clearly see that minimizing the objective function (\ref{eq:objective}) leads to minimizing the average estimated loss (\ref{eq:avg_est_loss}) as mentioned in the Remark of Theorem \ref{eq:main_thm}. Secondly, we observe that as the model complexity grows, namely, as the model becomes larger and epoch increases, the training objective and the average estimated loss both decreases. Thirdly, however, as shown by Theorem \ref{eq:main_thm}, the average denoising loss (\ie, the true BER) clearly shows the bias-variance tradeoff similar to the test error curve in supervised learning. In fact, the concentration property originally asserted in \citep{MooMinLeeYoo16} holds only for simpler model like $40\times 4$ and no longer holds for larger models, as can be predicted by the variance term in Theorem \ref{thm:uniform_conv}. But, the variance term does not seem to grow exponentially in $L$ and $N$ in practice, suggesting that the bound in Theorem \ref{thm:uniform_conv} can be improved.  

Since we never are able to evaluate the true denoising loss (\ref{eq:ave_true_loss}) in practice, we propose to use a small number of validation images that resemble the characteristics of the test images for selecting appropriate hyperparameters for the test images. 
Hence, we chose 5 additional images of size $512\times512$ for the validation set, namely, Boat, Man, Couple, Hamilton (Halftone) and Scan (Scanned text)\footnote{The images are shown in the Supplementary Material.}, and corrupted them with BSC ($\delta=0.1$). Figure \ref{fig:varying_k} shows the effect of varying $\ell$ (with fixed epoch of 7) on the average true and estimated BER of each model on the validation set. Similarly as in Figure \ref{fig:varying_epoch}, we again observe that the gap between the true denoising loss and the esimated loss (\ie, the variance term) increases as the model complexity grows, and the concentration happens only for $40\times 4$ model for reasonable size of $\ell$'s. We also see that BER is quite robust to $\ell$ when $\ell$ is sufficiently large. 

\begin{table}[h]\label{table:image}\vspace{-0.2in}
    \caption{The BER results on the test set ($\delta=0.1$)}
\smallskip\noindent
\resizebox{\linewidth}{!}{%
     \begin{tabular}{| c|| c| c|c|c|c|}
    \hline
     Models & Einstein & Lena & Barbara & C.man  & Shannon\\ \hline
        \hline
       \multirow{4}{*}{}    
                         New& \textbf{0.336 ($\mathbf{k=112}$)} & \textbf{0.179 ($k=112$)} &  \textbf{0.294 ($k=112$)}&  \textbf{0.192 ($k=112$)}  & \textbf{0.299 ($k=112$)} \\ \cline{1-6}
                         \hline
                         Original & 0.404 ($k=36$) & 0.403 ($k=38$) &  0.457 ($k=27$) &  0.268 ($k=35$)  & 0.402 ($k=35$) \\ \cline{1-6}
                         \hline\hline
                         Best (cf.)& 0.289 ($k=180$) & 0.171 ($k=40$) &  0.286 ($k=40$)&  0.190 ($k=40$) & 0.292 ($k=84$) \\ \cline{1-6}
                         \hline
    \end{tabular}}
\end{table}\vspace{-.05in}
Table 1 summarizes the BER results (relative to $\delta$) on the 5 test images. The first row (``New'') is for the Neural DUDE that uses hyperparameters selected from the validation set, and the second row (``Original'') is the results from \citep[Table 1]{MooMinLeeYoo16}. The best hyperparameters selected for ``New'' were $128\times 12$ model, $\ell=15 \ (\ie, k=112)$ and the number of epoch equal to $5$. Note while the hyperparameters are selected from the valiation set (with clean and noisy pairs), the network parameters are still learned with pseudo-labels derived from the \emph{test noisy} images, and \emph{no} supervised dataset is used for training Neural DUDE. For ``Original'', the hyperparameters were $40\times4$ model, different $k$ values for each image, and the number of epochs was 10.


From the table, we find that using 2-D contexts and network with much larger number of nodes, layers and $k$ values, determined by the valiation set, can achieve much smaller BER compared to the original results in \citep{MooMinLeeYoo16}. On average, the relative BER reduction of ``New'' compared to ``Original'' is about 32.8\%, which is significant. For comparison, the third row of Table 1 (``Best(cf.)'') is the result of the models of which hyperparameters are selected to obtain the best result for each test image (\ie, a genie-aided). We note that the result of ``New'' is not far from ``Best(cf.)'', just relatively 5.7\% larger, which concludes the validity of using a small validation set to optimize the bias-variance tradeoff predicted by Theorem \ref{eq:main_thm} for Neural DUDE.

\section{Concluding Remarks and Future Work}\label{sec:conclusion}

We gave a theoretical justification of the recently proposed Neural DUDE for univeral discrete denoising. We made a unique connection of the statistical learning theory to the denoising problem and obtained a denoising loss bound for the schemes designed to minimize the empirical estimated loss (\ref{eq:avg_est_loss}). The resulting bound shows the existence of the bias-variance tradeoff simliar to the standard supervised learning. The key result for obtaining such bound was to show the uniform concentration of the average estimated loss on the average true denoising loss in Theorem \ref{thm:uniform_conv}. Our theory suggests using separate validation set to select the hyperparameters for the neural network. In our experiments, we show that such procedure can achieve much lower BER than the original results with hand-picked hyperparameters. For future work, we plan to develop similar theory for continuous-valued signal case, \citep{ChaMoo18}. Practically, systematically applying Neural DUDE using validation set for the real DNA denoising task would be promising as the origianl DUDE already achieved competitive performance with the state-of-the-arts \citep{LeeMooYooWei16}. Furthermore, developing new regularization techniques for the denoising problem, \ie, slightly increasing the bias and decreasing the variance to minimize the denoising loss, would be another direction to pursue.

%% file: appendix.tex

\newpage
\section{Detailed proofs}
\subsection{Proof of Proposition 3}
For the notational brevity, we denote
\be
(\star) \triangleq \frac{1}{n-2k}\sum_{i=k+1}^{n-k}[\Ellb(Z_i,s_k(\Cb_i,\cdot))-\Lb(x_i,s_k(\Cb_i,Z_j))]\nonumber.
\ee
Then, for $\epsilon>0$, we have 
\begin{align}
&\text{Pr}\Big(\max_{s_k\in\mcS_k}(\star)>\epsilon\Big)\\
\leq& \sum_{s_k\in\mcS_k}\text{Pr}\Big((\star)>\epsilon\Big)
\leq|\mcS_k|(k+1)\exp\Big(-\frac{2(n-2k)\epsilon^2}{(k+1)C_{\max}^2}\Big)\label{eq:prop_2}\\
\leq&|\mcS|^{|\mcZ|^{2k}}(k+1)\exp\Big(-\frac{2(n-2k)\epsilon^2}{(k+1)C_{\max}^2}\Big)\label{eq:prop_3}
\end{align}
in which $C_{\max}\triangleq \max_{z,s}|\Ellb(z,s)|+\max_{x,\hat{x}}|\Lb(x,\hat{x})|$, the first inequality in (\ref{eq:prop_2}) follows from the union bound, the second inequality in (\ref{eq:prop_2}) follows from \citep[Lemma 2]{sdude}
, and (\ref{eq:prop_3}) follows from computing the size of $|\mcS_k|$. Now, by equating (\ref{eq:prop_3}) with $\delta$ and solving for $\epsilon$, we obtain
\be
\epsilon = \sqrt{\frac{(k+1)C_{\max}^2}{2(n-2k)} \Big(\log\big(\frac{k+1}{\delta}\big)+|\mcZ|^{2k}\log\big(\frac{|\mcS|}{\delta}\big) \Big) }.
\ee Thus, we have proven the propostion. \qed

\subsection{Proof of Lemma 1}
The lemma follows from the McDiarmid's inequality \citep{McDiarmid89}. That is, since we assume in the universal setting that the source sequence $x^n$ is an \emph{individual sequence}, we highlight that the randomness in $G_n$ is determined by the random variables $(Z_1,\ldots,Z_n)$ as follows:
\begin{align}
G_n = \sup_{\mathbf{w}\in\mathcal{W}_{k}^{L,N}}\tilde{\mathbf{R}}^\gamma_n[\mathbf{w}]\triangleq g(Z_1,\ldots,Z_n).\label{eq:def_Gn}
\end{align}
Then, by considering a different noisy sequence $(Z_1,\ldots,Z'_j,\ldots,Z_n)$ which is identical to $(Z_1,\ldots,Z_n)$ except for the $j$-th location replaced with $Z_j'$, the following bound on the difference holds :
\begin{align}
&|g(Z_1,\ldots,Z_j,\ldots,Z_n)-g(Z_1,\ldots,Z'_j,\ldots,Z_n)|\nonumber\\
=&\bigg|\sup_{\mathbf{w}\in\mathcal{W}_{k}^{L,N}}\Big(\frac{1}{n}\sum_{i=1}^n\tilde{r}^\gamma_i[\mathbf{w}]\Big)\nonumber\\
&-\sup_{\mathbf{w}\in\mathcal{W}_{k}^{L,N}}\Big(\frac{1}{n}\sum_{i=1}^n\tilde{r}^\gamma_i[\mathbf{w}]+\frac{1}{n}\sum_{i=j-k}^{j+k}\big\{\tilde{r}'^{\gamma}_i[\wb]-\tilde{r}^\gamma_i[\wb]\big\}\Big)\bigg|\label{eq:b_diff_1}\\
\leq& \bigg|\sup_{\mathbf{w}\in\mathcal{W}_{k}^{L,N}}\Big(\frac{1}{n}\sum_{i=j-k}^{j+k}\big\{\tilde{r}'^{\gamma}_i[\wb]-\tilde{r}^{\gamma}_i[\wb]\big\}\Big)\bigg|\label{eq:b_diff_4}\\
\leq& \frac{2(2k+1)}{n}C_{\max}\label{eq:b_diff_3},
\end{align}
in which $\tilde{r}'^{\gamma}_i[\wb]$ in (\ref{eq:b_diff_1}) stands for the interpolated per-symbol regret that is affected by $Z'_j$, (\ref{eq:b_diff_1}) follows from the fact that $\tilde{r}^\gamma_i[\wb]$ in [Manuscript, Eq.(12)] is only affected by the $(2k+1)$-tuple, $Z_{i-k}^{i+k}$, the (\ref{eq:b_diff_4}) follows from separately applying the supremem for the second term, and (\ref{eq:b_diff_3}) follows from the fact that for all $\mathbf{w}\in\mathcal{W}_{k}^{L,N}$,
\begin{align}
&\Big|\frac{1}{n}\sum_{i=j-k}^{j+k}\big\{\tilde{r}'^\gamma_i[\wb]-\tilde{r}^\gamma_i[\wb]\big\}\Big|\nonumber\\
\leq& \frac{1}{n}\sum_{i=j-k}^{j+k}\{|\tilde{r}'^\gamma_i[\wb]|+|\tilde{r}^\gamma_i[\wb]|\}\leq \frac{2(2k+1)}{n}C_{\max}.
\end{align}
Now, since the noisy observations $(Z_1,\ldots,Z_n)$ are independent random variables given the underlying clean source $(x_1,\ldots,x_n)$ from the memoryless channel assumption, we can directly apply the McDiarmid's inequality and obtain the lemma. \qed

\subsection{Proof of Lemma 2}

First, by referring to the notations [Manuscript, Eq.(4)] and [Manuscript, Eq.(12)], we have
\be
\tilde{r}^\gamma_i[\wb]= r_i[\wb]+ \Delta r_i^\gamma[\wb]\label{eq:delta_r_w_def}
\ee
by defining $\Delta_i^\gamma[\wb]\triangleq\Delta r_{(x_i,Z_i)}^\gamma(\mathbf{p}^k(\wb,\Cb_i))$. Then, we have the following inequalities:
\begin{align}
&\mathbb{E}(G_n)=
\mathbb{E}\bigg(\sup_{\mathbf{w}\in\mathcal{W}_k^{L,N}}\frac{1}{n}\sum_{i=1}^n\tilde{r}^\gamma_i[\wb]\bigg)\\
=&\mathbb{E}\bigg(\sup_{\mathbf{w}\in\mathcal{W}_k^{L,N}}\Big[\frac{1}{n}\sum_{i=1}^n\tilde{r}^\gamma_i[\wb]-\mathbb{E}\Big(\frac{1}{n}\sum_{i=1}^n\tilde{r}^\gamma_i[\wb]\Big)\nonumber\\
&\ \ \ +\mathbb{E}\Big(\frac{1}{n}\sum_{i=1}^n\Delta r^\gamma_i[\wb]\Big)\Big]\bigg)\label{eq:rc_1}\\
\leq&\mathbb{E}\bigg(\sup_{\mathbf{w}\in\mathcal{W}_k^{L,N}}\Big[\frac{1}{n}\sum_{i=1}^n\tilde{r}^\gamma_i[\wb]-\mathbb{E}\Big(\frac{1}{n}\sum_{i=1}^n\tilde{r}^\gamma_i[\wb]\Big)\Big]\bigg)\nonumber\\
& +2C_{\max}\gamma|\mcS|^2,\label{eq:rc_1_1}
\end{align}
in which (\ref{eq:rc_1}) follows from (\ref{eq:delta_r_w_def}) and the fact 
\begin{align}
&\mathbb{E}(r_i[\wb])\nonumber\\=& \mathbb{E}\big(\mathbf{L}(Z_i,s_k[\wb](\Cb_i,\cdot))-\Lb(x_i,s_k[\wb](\Cb_i,Z_i))\big) \nonumber\\
=&\mathbb{E}\Big(\mathbb{E}\big(\mathbf{L}(Z_i,s_k[\wb](\Cb_i,\cdot))-\Lb(x_i,s_k[\wb](\Cb_i,Z_i))\Big|Z^{\backslash i}\big)\Big)\nonumber\\
=&0\label{eq:rc_4}
\end{align}
for all $i$, since (\ref{eq:rc_4}) is from the fact that $\Ellb(Z,s)$ is an unbiased estimate of $\mathbb{E}\Lb(x,s(Z))$ as defined in [Manuscript, Section 2]. Thus, $\mathbb{E}(\tilde{r}^\gamma_i[\wb])=\mathbb{E}(\Delta r^\gamma_i[\wb])$ for all $i$, and we have (\ref{eq:rc_1}). Furthermore, (\ref{eq:rc_1_1}) is from the bound 
\be
\mathbb{E}(\Delta r^\gamma_i[\wb])\leq 2C_{\max}\gamma|\mcS|^2, \ \ \text{for all } i
\ee
since $2C_{\max}\gamma$ is the crude upper bound on the maximum contribution to the expectation at each decision boundary, and there are no more than $|\mcS|^2$ decision boundaries in the simplex. 

Now, similarly as in the arguments for the generalization bound and Rademacher complexity in learning theory \citep{cs229T}, we introduce a \emph{ghost} noisy observation sequence $(Z_1^*,\ldots,Z_n^*)$, which is an independent realization of the noisy observations given the same underlying clean source sequence $(x_1,\ldots,x_n)$. 
By defining $\tilde{r}^{\gamma*}_i[\wb]$ as the interporlated regret function computed with $(x_i,Z_i^*)$, we can continue the inequalities for the expectation term in (\ref{eq:rc_1_1}) as follows:
\begin{align}
&\mathbb{E}\Big(\sup_{\mathbf{w}\in\mathcal{W}_k^{L,N}}\Big[\frac{1}{n}\sum_{i=1}^n\tilde{r}^\gamma_i[\wb]-\mathbb{E}\Big(\frac{1}{n}\sum_{i=1}^n\tilde{r}^\gamma_i[\wb]\Big)\Big]\Big)\nonumber\\
=&\mathbb{E}\Big(\sup_{\mathbf{w}\in\mathcal{W}_k^{L,N}}\Big[\frac{1}{n}\sum_{i=1}^n\tilde{r}^\gamma_i[\wb]-\mathbb{E}\Big(\frac{1}{n}\sum_{i=1}^n\tilde{r}^{\gamma*}_i[\wb]\Big)\Big]\Big)\label{eq:rc_2}
\end{align}
\begin{align}
=&\mathbb{E}\Big(\sup_{\mathbf{w}\in\mathcal{W}_k^{L,N}}\mathbb{E}\Big(\frac{1}{n}\sum_{i=1}^n\tilde{r}^\gamma_i[\wb]-\frac{1}{n}\sum_{i=1}^n\tilde{r}^{\gamma*}_i[\wb]\Big|Z^n\Big)\Big)\label{eq:rc_3}\\
\leq&\mathbb{E}\Big(\mathbb{E}\Big(\sup_{\mathbf{w}\in\mathcal{W}_k^{L,N}}\Big[\frac{1}{n}\sum_{i=1}^n\tilde{r}^\gamma_i[\wb]-\frac{1}{n}\sum_{i=1}^n\tilde{r}^{\gamma*}_i[\wb]\Big]\Big|Z^n\Big)\Big)\label{eq:rc_5}\\
=&\mathbb{E}\Big(\sup_{\mathbf{w}\in\mathcal{W}_k^{L,N}}\frac{1}{n}\sum_{i=1}^n\big[\tilde{r}^\gamma_i[\wb]-\tilde{r}^{\gamma*}_i[\wb]\big]\Big)\label{eq:rc_6}
\end{align}
in which (\ref{eq:rc_2}) is from the fact that both $\frac{1}{n}\sum_{i=1}^n\tilde{r}^\gamma_i[\wb]$ and $\frac{1}{n}\sum_{i=1}^n\tilde{r}^{\gamma*}_i[\wb]$ follow the identical distribution, (\ref{eq:rc_3}) follows from the fact that $(Z_1,\ldots,Z_n)$ and $(Z_1^*,\ldots,Z_n^*)$ are independent, (\ref{eq:rc_5}) follows from pushing the supremum inside the expectation, and (\ref{eq:rc_6}) follows from carrying out the iterated conditional expectation.

Now, we introduce the Rademacher variables $\sigma_1,\ldots, \sigma_n$ that are independent of $(Z_1,\ldots,Z_n)$ and $(Z^*_1,\ldots,Z^*_n)$, and each $\sigma_i$ is i.i.d and uniform over $\{+1,-1\}$. Then, continuing the inequalities from (\ref{eq:rc_6}) yields
\begin{align}
&\text{Eq.} (\ref{eq:rc_6})\\
=&\mathbb{E}\Big(\sup_{\mathbf{w}\in\mathcal{W}_k^{L,N}}\frac{1}{n}\sum_{i=1}^n\sigma_i\big[\tilde{r}^\gamma_i[\wb]-\tilde{r}^{\gamma*}_i[\wb]\big]\Big)\label{eq:rc_7}\\
\leq&\mathbb{E}\Big(\sup_{\mathbf{w}\in\mathcal{W}_k^{L,N}}\frac{1}{n}\sum_{i=1}^n\sigma_i\tilde{r}^\gamma_i[\wb]\nonumber\\
&\ \ \ +\sup_{\mathbf{w}\in\mathcal{W}_k^{L,N}}\frac{1}{n}\sum_{i=1}^n(-\sigma_i)\tilde{r}^{\gamma*}_i[\wb]\Big)\label{eq:rc_8}\\
=&2\mathbb{E}\Big(\sup_{\mathbf{w}\in\mathcal{W}_k^{L,N}}\frac{1}{n}\sum_{i=1}^n\sigma_i\tilde{r}^\gamma_i[\wb]\Big)\label{eq:rc_9}\\
=&2R_n(\mathcal{A})\label{eq:rc_10}
\end{align}
in which (\ref{eq:rc_7}) follows from the fact that the distribution of $\tilde{r}^\gamma_i[\wb]-\tilde{r}^{\gamma*}_i[\wb]$ is symmetric around 0 and multiplying $\sigma_i$ does not change the distribution, (\ref{eq:rc_8}) holds by the inequality $\sup_\wb \{a[\wb]-b[\wb]\}\leq \sup_{\wb}a[\wb]+\sup_\wb\{-b[\wb]\}$, (\ref{eq:rc_9}) follows from the linearity of expectation and the fact that $\sigma_i$ and $-\sigma_i$ have the same distribution, and (\ref{eq:rc_10}) is from the definition of the Rademacher complexity [Manuscript, Eq.(19)]. Now, by combining (\ref{eq:rc_10}) and (\ref{eq:rc_1_1}), the lemma is proven. \qed

\subsection{Proof of Lemma 3}

First, note that the definition of the Rademacher complexity in [Manuscript, Eq.(19)] is slightly different from the ordinary definition in \citep[Eq.(218)]{cs229T} in that the summands $\tilde{r}^\gamma_i[\wb]$'s are not independent due to the overlapping contexts. However, we can still utilize the general tools of Rademacher complexity to obtain the bound [Manuscript, Eq.(20)] of the lemma. 

We will show the bound with consecutive compositions of the layers. First, for $\ell\geq2$, we denote $\mathbf{w}_{\ell,m}\in\mathbb{R}^{n_{\ell-1}}$ as the weight parameter vector associated with the $m$-th node in the $\ell$-th layer. For $\ell=1$, we have $\mathbf{w}_{1,m}\in\mathbb{R}^{2|\mcZ|k}$ as the input to the network, $\Cb_i=(Z_{i-k}^{i-1},Z_{i+1}^{i+k})$, has dimension $2|\mcZ|k$, since each noisy symbol is one-hot encoded with dimension $|\mcZ|$.

 From our assumption, we have $\|\wb_{\ell,m}\|_2\leq B$ for each $\ell$ and $m$.
 Furthermore, $\|\Cb_i\|_2\leq \sqrt{2k}$ \emph{a.s.} due to the one-hot encoding of the noisy symbols. Since we are using the ReLU function, $f(x)=\max\{0,x\}$, as the nonlinearity in the network, we first consider the Rademacher complexity of the function class $\mathcal{F}=\{(x,z_0,\cb)\mapsto \mathbf{w}^\top\cb:\|\wb\|_2\leq B\}$ in which $\cb=(z_{-k}^{-1},z_1^k)\in\Cb[k]$. That is, we have
 \begin{align}
&\mathcal{R}_n(\mathcal{F})\nonumber\\
=&\frac{1}{n}\mathbb{E}\Bigg(\sup_{\|\mathbf{w}\|_2\leq B} \sum_{i=1}^n\sigma_i(\mathbf{w}^\top\Cb_i)   \Bigg)\nonumber\\
\leq & \frac{ B}{n}\mathbb{E}\bigg( \Big\|\sum_{i=1}^n\sigma_i\Cb_i \Big\|_2 \bigg)\label{eq:rc1_bound1_1}\\
\leq& \frac{ B}{n}\sqrt{\mathbb{E}\bigg( \Big\|\sum_{i=1}^n\sigma_i\Cb_i \Big\|_2^2 \bigg)}\label{eq:rc1_bound1}\\
\leq& \frac{B}{n}\sqrt{\mathbb{E}\bigg( \sum_{i=1}^n\|\sigma_i\Cb_i \|_2^2 \bigg)}\label{eq:rc1_bound2_1}\\
=&\frac{ B}{n}\sqrt{\mathbb{E}\bigg( \sum_{i=1}^n\|\Cb_i \|_2^2 \bigg)}\label{eq:rc1_bound2}\\
\leq& \frac{B}{n}\sqrt{2kn} =B\sqrt{\frac{2k}{n}}, \label{eq:rc1_bound3}
 \end{align}
in which (\ref{eq:rc1_bound1_1}) follows from applying Cauchy-Schwartz inequality to $\mathbf{w}$ and $\sum_{i=1}^n\sigma_i\Cb_i$, (\ref{eq:rc1_bound1}) follows from the concavity of $\sqrt{\cdot}$, (\ref{eq:rc1_bound2_1}) follows from the independence of $\sigma_i$, (\ref{eq:rc1_bound2}) follows from the fact that $\sigma_i$ does not affect the norm, and (\ref{eq:rc1_bound3}) follows the bound on $\|\Cb_i\|_2$. Note that the independence of $\Cb_i$ is \emph{not} required for the bounds. 

Now, from the Lipschitz composition property of the Rademacher complexity \citep[Corollary 3.17]{LedTal91}, we upper bound the Rademacher complexity of the hidden node in the first layer as
\begin{align}
&\mathcal{R}_n\big(\{(x,z_0,\cb)\mapsto f(\mathbf{w}^\top\mathbf{c}): \|\mathbf{w}\|_2\leq  B\}\big)\nonumber\\
\leq&B\sqrt{\frac{2k}{n}}, \label{eq:lem3_eq1}
\end{align}
since $f(x)=\max\{0,x\}$ is 1-Lipschitz continuous.

Continuing to the second layer, by denoting $\mathbf{h}_1=[f(\wb_{1,1}^\top\cb),\ldots,f(\wb_{1,n_1}^\top\cb)]^\top\in\mathbb{R}^{n_1}$ as the vector of the first hidden layer node values for the input $\cb\in\Cb[k]$, the value of the $m$-th node in the second layer is $f(\mathbf{w}_{2,m}^\top\mathbf{h}_1)$ with $\|\mathbf{w}_{2,m}\|_2\leq B$. Then, by following the similar argument as in \citep[Theorem 43]{cs229T}, we have 
\begin{align}
&\mathcal{R}_n\big(\{(x,z_0,\cb)\mapsto f(\mathbf{w}_{2,m}^\top\mathbf{h}_1): \|\mathbf{w}_{2,m}\|_2\leq B\}\big)\nonumber\\
\leq&(2B^2)\sqrt{\frac{2kn_1}{n}},
\end{align}
again by the Lipschitz continuity of $f(x)$ and the fact that $f(0)=0$.

By continuing above argument, we can see that the Rademacher complexity of the $s$-th node in the final output layer, $\mathbf{o}_s$, before the softmax function can be bounded by 
\begin{align}
(2B)^{L+1}\sqrt{\frac{k\big(\prod_{\ell=1}^{L}n_\ell\big)|\mcS|}{2n}}.
\end{align}
Since we can show that the softmax function $\{\mathbf{o}\mapsto \mathbf{p}=\text{softmax}(\mathbf{o}_1,\ldots,\mathbf{o}_{|\mcS|})$ is also 1-Lipschitz \citep[Corollary 3]4{GaoPav17}, we have 
\be
&&\mathcal{R}_n\big(\{(x,z_0,\cb)\mapsto \mathbf{p}^k(\wb,\cb): \mathbf{w}\in\mathcal{W}_k^{L,N}\}\nonumber\\
&\leq& (2B)^{L+1}\sqrt{\frac{k\big(\prod_{\ell=1}^{L}n_\ell\big)|\mcS|}{2n}}.
\ee
Finally, since $\tilde{r}^\gamma_{(x,z)}(\mathbf{p})$ in [Manuscript, Eq.(12)] is designed to be $\frac{C_{\max}}{\gamma}$-Lipschitz as in [Manuscript, Eq.(14)], we can apply the Lipschitz composition property one more time and have the bound in the lemma. \ \qed

\section{Test and validation images}

\begin{figure*}[h]
    \centering
    \subfigure[Einstein]{\label{fig:n_dude_arch}
    \includegraphics[width=0.17\textwidth]{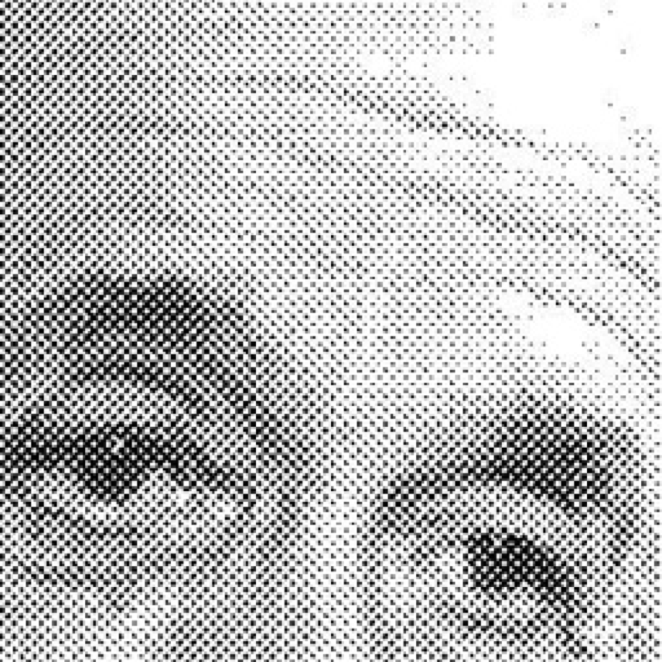}}
    \subfigure[Lena]{\label{fig:n_dude_arch}
    \includegraphics[width=0.17\textwidth]{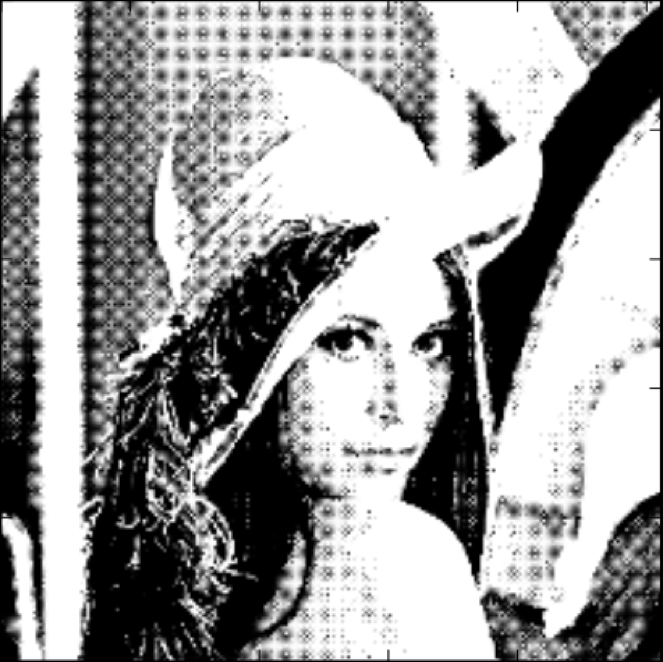}}
    \subfigure[Barbara]{\label{fig:n_dude_arch}
    \includegraphics[width=0.17\textwidth]{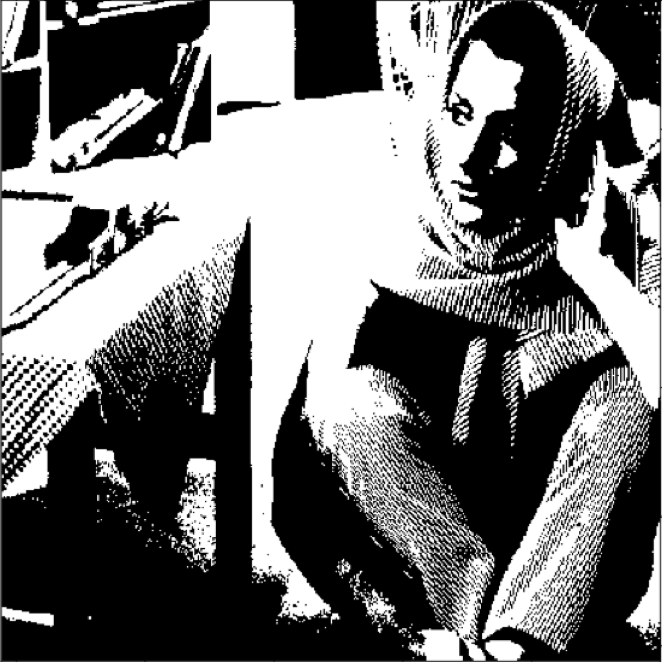}}
    \subfigure[C.man]{\label{fig:n_dude_arch}
    \includegraphics[width=0.17\textwidth]{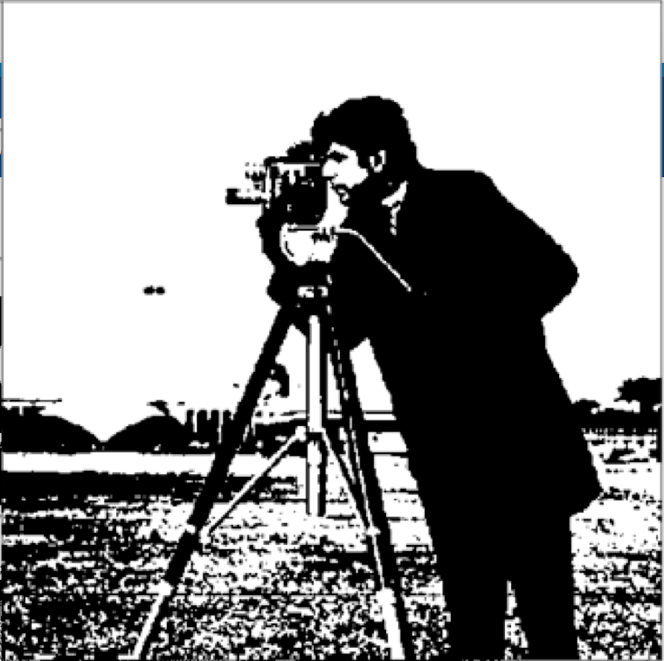}}
    \subfigure[Shannon]{\label{fig:n_dude_arch}
    \includegraphics[width=0.17\textwidth]{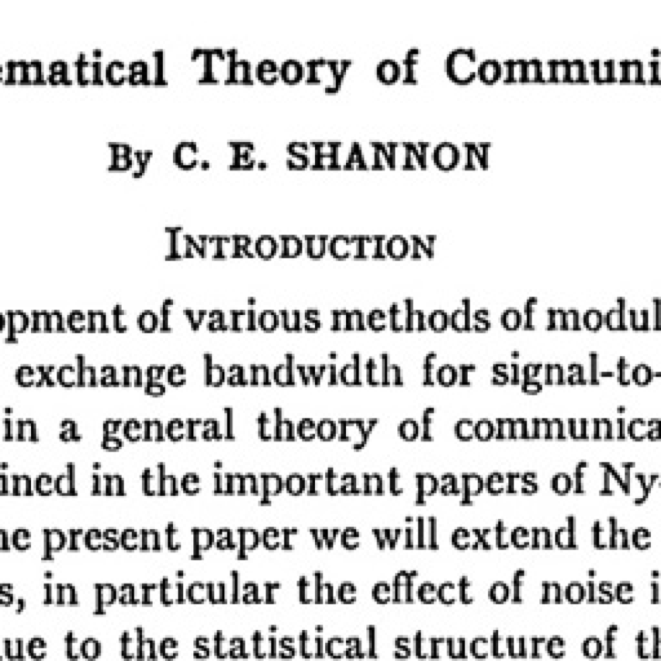}}
    \caption{5 \textbf{test} images used for experiments
    }
    \label{fig:test_images}
\end{figure*}

\begin{figure*}[h]
    \centering
    \subfigure[Hamilton]{\label{fig:n_dude_arch}
    \includegraphics[width=0.17\textwidth]{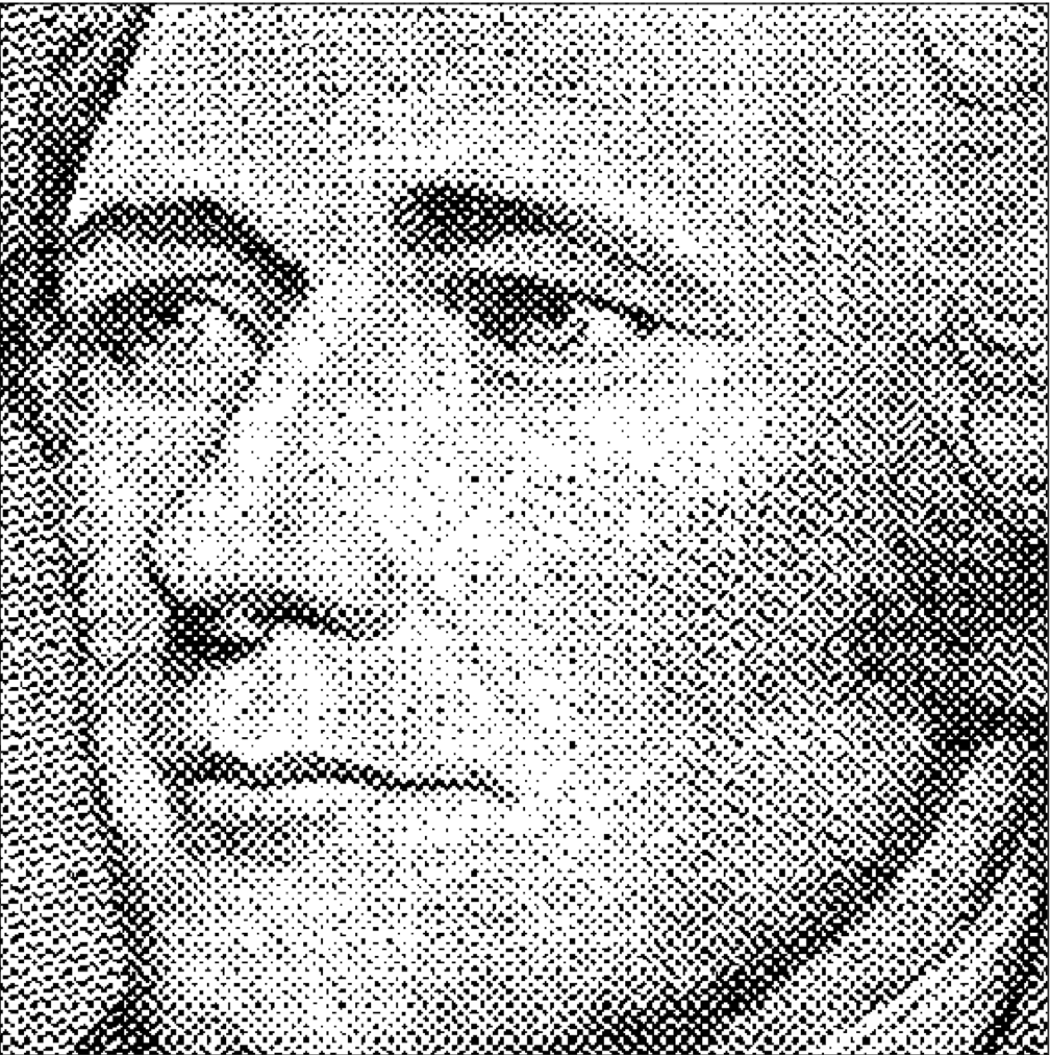}}
    \subfigure[Boat]{\label{fig:n_dude_arch}
    \includegraphics[width=0.17\textwidth]{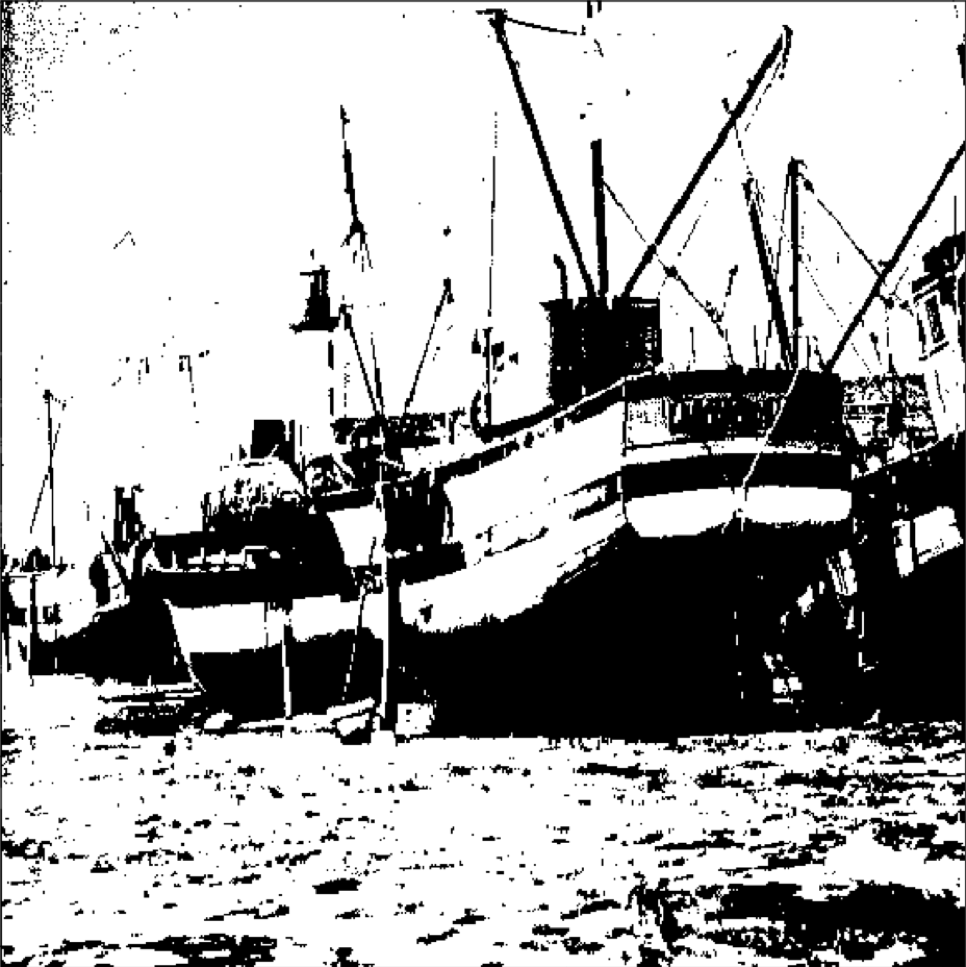}}
    \subfigure[Man]{\label{fig:n_dude_arch}
    \includegraphics[width=0.17\textwidth]{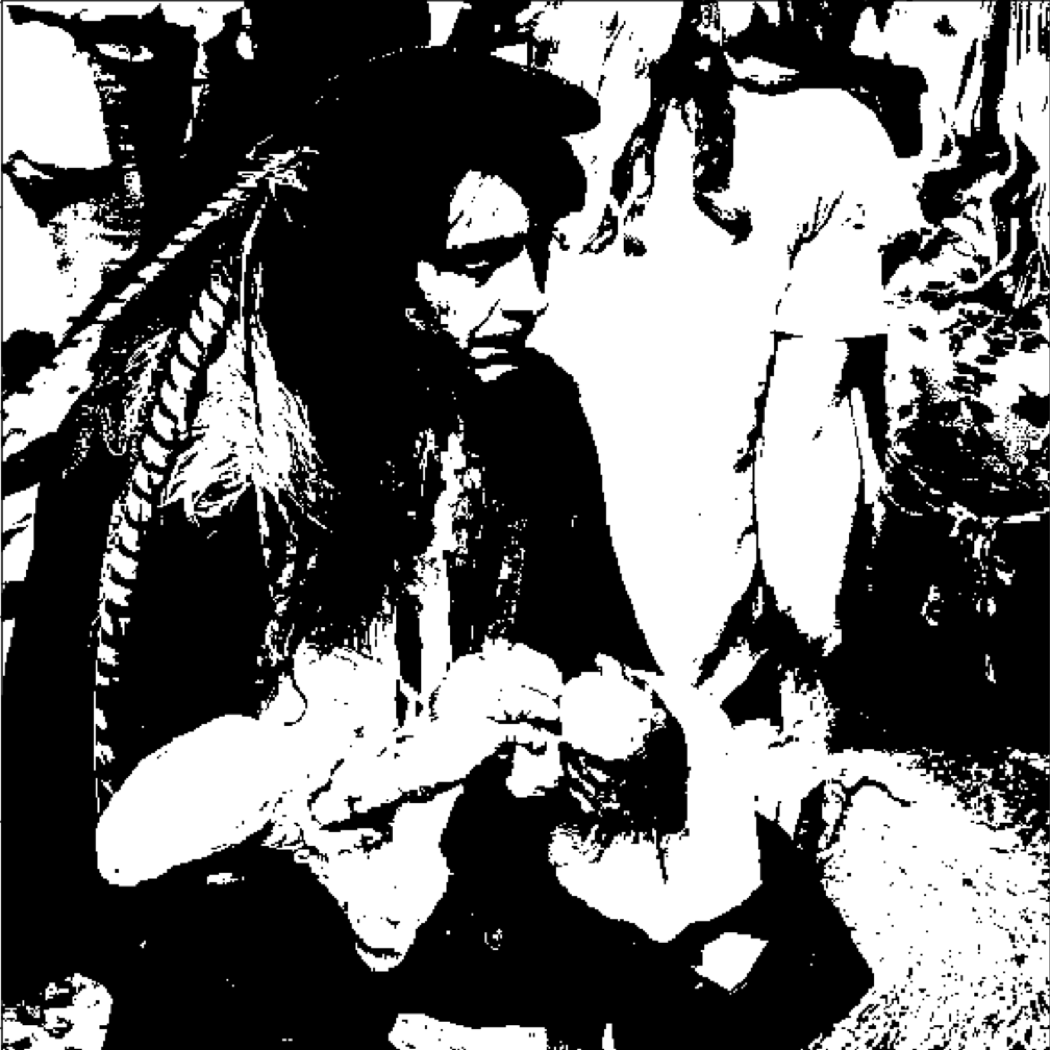}}
    \subfigure[Couple]{\label{fig:n_dude_arch}
    \includegraphics[width=0.17\textwidth]{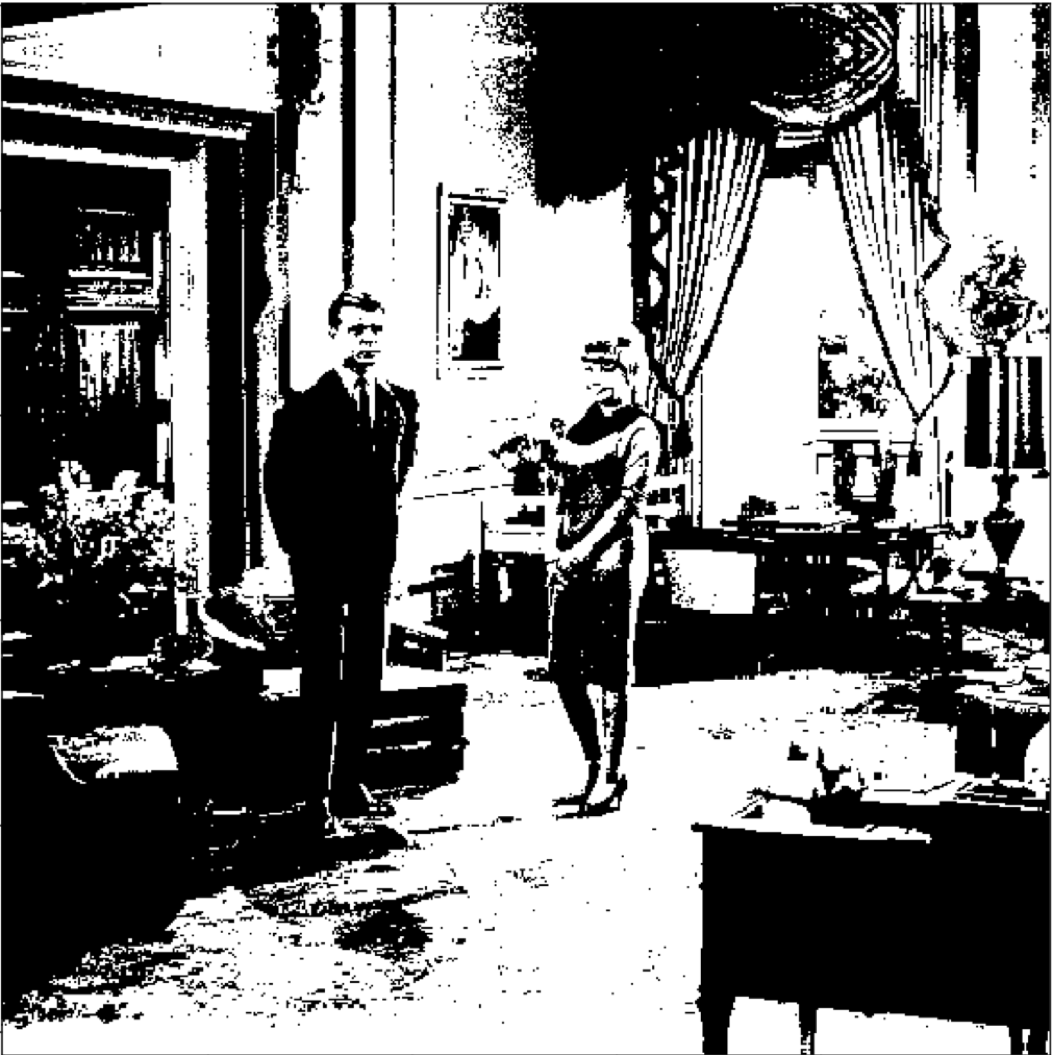}}
    \subfigure[Scan]{\label{fig:n_dude_arch}
    \includegraphics[width=0.17\textwidth]{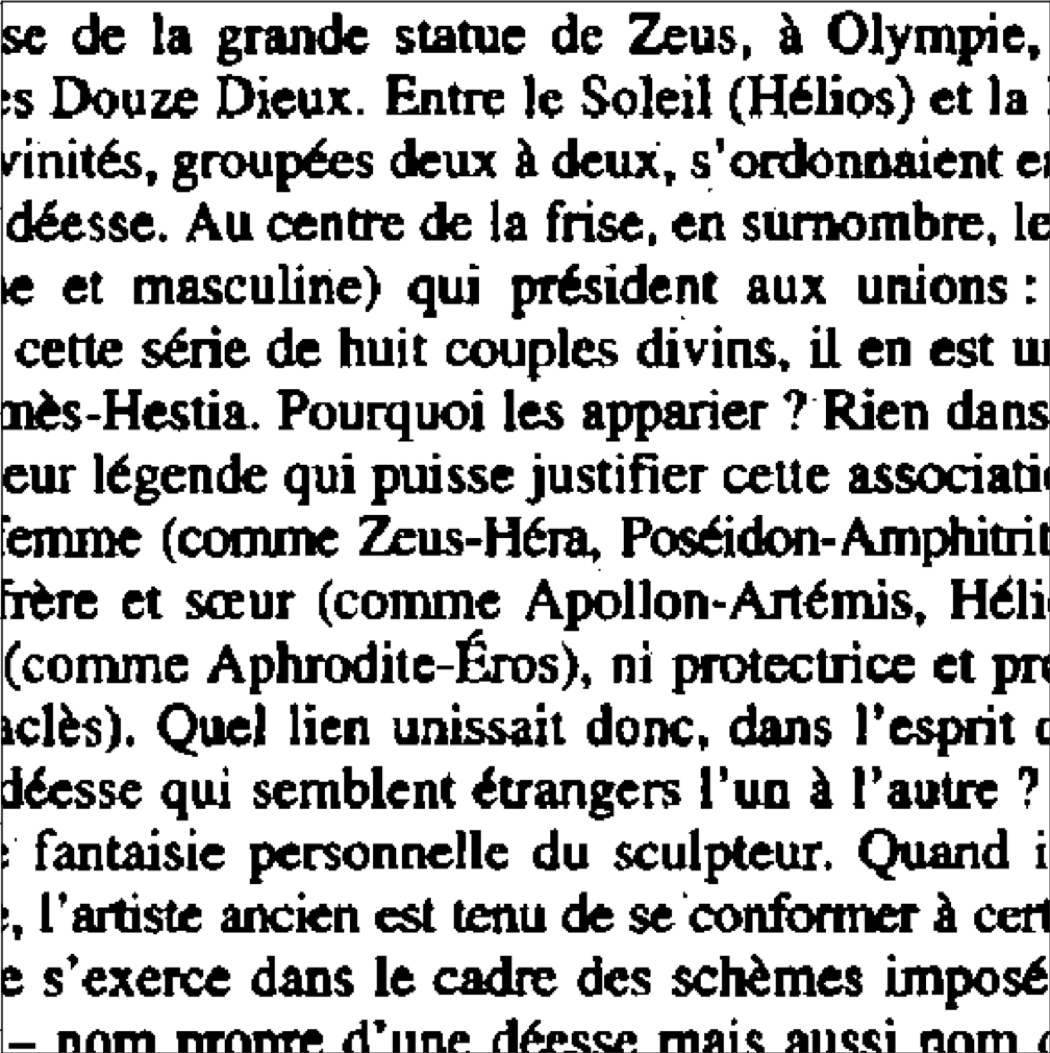}}
    \caption{5 \textbf{validation} images used for experiments
    }
    \label{fig:val_images}
\end{figure*}

Here, we show the test and validation images that we used for the experiments in Section 5. Figure \ref{fig:test_images} shows the 5 test images and Figure \ref{fig:val_images} shows the 5 validation images. As can be seen in the figures, the validation images are certainly different from the test images, but possess similar visual characteristics with the test images.

